\documentclass{tlp}

\usepackage{srcltx}
\usepackage{braket}
\usepackage{url}

\usepackage[colorlinks]{hyperref}
\hypersetup{
	bookmarksnumbered,
	pdfstartview={FitH},
	citecolor={blue},
	linkcolor={red},
	urlcolor={magenta},
	pdfhighlight={/O},
	pdfpagemode={UseOutlines},
	pdfauthor={Martin Slota and Joao Leite},
	pdftitle={Towards Closed World Reasoning in Dynamic Open Worlds (Extended Version)},
	pdfkeywords={belief change, belief update, hybrid knowledge bases, ontologies, rules, description logics, answer set programming, semantic web}
}

\newtheorem{theorem}{Theorem}[section]
\newtheorem{proposition}[theorem]{Proposition}
\newtheorem{lemma}[theorem]{Lemma}
\newtheorem{corollary}[theorem]{Corollary}
\newtheorem{definition}[theorem]{Definition}

\newtheorem{example}[theorem]{Example}

\newenvironment{proposition*}[1]{\par \noindent \textit{Proposition \ref{#1}.} \it}{}
\newenvironment{corollary*}[1]{\par \noindent \textit{Corollary \ref{#1}.} \it}{}

\newenvironment{itemize*}%
  {\begin{itemize}%
    \setlength{\itemsep}{0pt}%
    \setlength{\parskip}{0pt}}%
  {\end{itemize}}
\newenvironment{enumerate*}%
  {\begin{enumerate}%
    \setlength{\itemsep}{0pt}%
    \setlength{\parskip}{0pt}}%
  {\end{enumerate}}

\newcommand{\qedhere}{\qquad \mathproofbox}

\usepackage{extpfeil}

\newcommand{\br}[1]{\left( #1 \right)}
\newcommand{\an}[1]{\left\langle #1 \right\rangle}

\newextarrow{\myxlongequal}{3300}{\Relbar\Relbar\Relbar}
\newcommand{\eqbyeq}[1][]{\myxlongequal[\phantom{\text{(11)}}]{#1}}
\newcommand{\eqbylemma}[1][]{\myxlongequal[\phantom{\text{\,Lemma 1.11\,}}]{#1}}

\newcommand{\mlthen}{\Rightarrow}

\newcommand{\mlequiv}{\Longleftrightarrow}

\newcommand{\mlequivbylemma}[1]{\xLeftrightarrow[\phantom{\text{~Lemma 1.11~}}]{#1}}

\newcommand{\nat}{\mathbb{N}}

\newcommand{\lang}{\mathcal{L}}
\newcommand{\lvar}{\mathbf{V}}
\newcommand{\lcon}{\mathbf{C}}

\newcommand{\lpre}{\mathbf{P}}

\newcommand{\ltrue}{\mathsf{true}}
\newcommand{\lfalse}{\mathsf{false}}

\newcommand{\lif}{\subset}
\newcommand{\lthen}{\supset}
\newcommand{\lequiv}{\equiv}

\newcommand{\mymodels}{\mathrel\mid\joinrel=}
\newcommand{\ent}{\mymodels}
\newcommand{\nent}{\not\ent}

\newcommand{\smod}{\mathsf{mod}}

\newcommand{\mk}[1][\,]{\mathbf{K}#1}
\newcommand{\mnot}[1][\,]{\mathbf{not}#1}
\newcommand{\mstr}[1][I, M, N]{\left\langle #1 \right\rangle}

\newcommand{\restr}[2][\lpre', \lcon']{#2^{\left[#1\right]}}

\newcommand{\lconf}[1][\phi]{\lcon^{[#1]}}
\newcommand{\lpref}[1][\phi]{\lpre^{[#1]}}

\newcommand{\foint}{\mathcal{I}}
\newcommand{\mint}{\mathcal{M}}

\newcommand{\kb}[1][K]{\mathcal{#1}}
\newcommand{\thr}[1][S]{\mathcal{#1}}

\newcommand{\ont}[1][O]{\mathcal{#1}}
\newcommand{\tbox}[1][T]{\mathcal{#1}}
\newcommand{\abox}[1][A]{\mathcal{#1}}

\newcommand{\lpnot}[1][\,]{\mathit{not}#1}
\newcommand{\lpif}{\leftarrow}

\newcommand{\defmod}{M}

\newcommand{\prog}[1][\defprog]{\mathcal{#1}}

\newcommand{\progmod}[1][\defprog]{{{\prog[#1]}^{\defmod}}}

\newcommand{\tkb}[1][K]{T_{\mathcal{#1}}}

\begin{document}

\title[Towards Closed World Reasoning in Dynamic Open Worlds]{Towards Closed World Reasoning \\ in Dynamic Open Worlds \\ (Extended Version)}

\author[M. Slota and J. Leite]
{MARTIN SLOTA$^{\footnotemark[1]}$ and JO{\~A}O LEITE \\
CENTRIA \& Departamento de Inform{\'a}tica \\
Universidade Nova de Lisboa \\
2829-516 Caparica, Portugal}

\renewcommand{\thefootnote}{\fnsymbol{footnote}}
\footnotetext[1]{Supported by FCT Scholarship SFRH/BD/38214/2007. Participation on conference supported by FLoC 2010 Student Travel Support and by APPIA Study Scholarship.}
\setcounter{footnote}{0}
\renewcommand{\thefootnote}{\arabic{footnote}}

\maketitle

\noindent \textbf{Note}: This is an extended version containing all proofs of the article published in \emph{Theory and Practice of Logic Programming}, \textbf{10} (4-6): 547 -- 564, July. \copyright 2010 Cambridge University Press.

\noindent \textbf{Changes on July 23, 2010}: some minor substitutions and additions to be in line with the TPLP version; full reference to the journal version added.

\bigskip

\begin{abstract}
The need for integration of ontologies with nonmonotonic rules has been gaining importance in a number of areas, such as the Semantic Web. A number of researchers addressed this problem by proposing a unified semantics for \emph{hybrid knowledge bases} composed of both an ontology (expressed in a fragment of first-order logic) and nonmonotonic rules. These semantics have matured over the years, but only provide solutions for the static case when knowledge does not need to evolve.

In this paper we take a first step towards addressing the dynamics of hybrid knowledge bases. We focus on knowledge updates and, considering the state of the art of belief update, ontology update and rule update, we show that current solutions are only partial and difficult to combine. Then we extend the existing work on ABox updates with rules, provide a semantics for such evolving hybrid knowledge bases and study its basic properties.

To the best of our knowledge, this is the first time that an update operator is proposed for hybrid knowledge bases.
\end{abstract}


\section{Introduction}

In this paper we address updates of hybrid knowledge bases composed of a Description Logic ontology and Logic Programming rules. We propose an operator to be used when a hybrid theory is updated by new observations of a changing world, examine its properties, and discuss open problems pointing to future research.

The Semantic Web was initiated almost a decade ago with an ambitious plan regarding the sharing of metadata and knowledge in the Web, enhanced with reasoning services for advanced new applications \cite{Berners-Lee2001}. Since then, the considerable amount of research devoted to this endeavour originated important foundational results and a deeper understanding of the issues involved, while identifying important conclusions regarding future developments, namely that:

\begin{enumerate}
	\item Ontologies are necessary and useful for knowledge representation in the Semantic Web. The formalisms developed, e.g. OWL, are powerful enough to capture existing modelling languages used in software engineering, and extend their capabilities. Ontologies are usually based on decidable, as well as tractable, fragments of Classical Logic, such as the Description Logics (DL) \cite{Baader2003}. They adopt the \emph{open world assumption} (OWA) i.e. they view a knowledge base, \emph{by assumption}, to be potentially incomplete, hence a proposition $p$ is false only if the knowledge base is inconsistent with $p$. This suits well the open nature of such systems where complete knowledge about the environment cannot be assumed.

	\item Rules are fundamental to overcome the limitations found in OWL. They enjoy formal, declarative and well-understood semantics, the \emph{stable model semantics} \cite{Gelfond1988} and its tractable approximation, the three-valued \emph{well-founded semantics} \cite{Gelder1991} being the most prominent and widely accepted. These semantics adopt the \emph{closed world assumption} (CWA) i.e. the knowledge base \emph{is assumed} to contain complete information. Consequently, a proposition $p$ is considered false whenever it is not entailed to be true. This type of negation is usually dubbed \emph{default negation} or \emph{weak negation}, to distinguish it from the \emph{classical negation} used in Classical Logic. Rules can naturally express assumptions, policies, preferences, norms and laws, and provide constructs which are more natural for software developers (as used in Relational Databases and Logic Programming).

	\item The open and dynamic character of the Semantic Web requires new knowledge based systems to be equipped with mechanisms to evolve.
\end{enumerate}

Indeed, the growing availability of information requires the support of dynamic data and application integration, automation and interoperation of business processes and problem-solving in various domains, to enforce correctness of decisions, and to allow traceability of the knowledge used and of the decisions taken. In these scenarios, ontologies provide the logical foundation of intelligent access and information integration, while rules are used to represent business policies, regulations and declarative guidelines about information, and mappings between different information sources.

Over the last decade, there have been many proposals for integrating DL based monotonic ontologies with nonmonotonic rules (see \cite{Hitzler2009} for a survey). Recently, in \cite{Motik2007},  Hybrid MKNF Knowledge Bases were introduced, allowing predicates to be defined concurrently in both an ontology and a set of rules, while enjoying several important properties. There is even a tractable variant based on the well-founded semantics that allows for a top-down querying procedure \cite{Alferes2009}, making the approach amenable to practical applications that need to deal with large ontologies.

But this only addresses part of the problem. The highly dynamic character of the Semantic Web calls for the development of ways to deal with updates of these hybrid knowledge bases composed of both rules and ontologies, and the inconsistencies that may arise. The dynamics of hybrid knowledge bases, to the best of our knowledge, has never been addressed before.

However, the problems associated with knowledge evolution have been extensively studied, over the years, by researchers in different research communities, namely in the context of Classical Logic, and in the context of Logic Programming. They proved to be extremely difficult to solve, and existing solutions, even within each community, are still subject of active debate as they do not seem adequate in all kinds of situations in which their application is desirable.

In the context of Classical Logic, the seminal work by Alchourr{\'{o}}n, G{\"{a}}rdenfors and Makinson (AGM) \cite{Alchourron1985} proposed a set of desirable properties of belief change operators, now called \emph{AGM postulates}. Subsequently, in \cite{Katsuno1991}, \emph{update} and \emph{revision} have been distinguished as two very related but ultimately different belief change operations. While revision deals with incorporating new information about a static world, update takes place when changes occurring in a dynamic world are recorded. The authors of \cite{Katsuno1991} formulated a separate set of postulates for updates. One of the specific update operators that satisfies these postulates is Winslett's minimal change update operator \cite{Winslett1990}. Though we believe that revision operators for hybrid knowledge bases pose an interesting and important research topic, in this paper we focus on update operators and do not tackle revision any further.

Further research showed that, in most cases, belief update operators cannot be directly applied to Description Logic ontologies. The existing work considers only ABox updates, allowing only for static acyclic TBoxes which are ``expanded'' before the update takes place \cite{Liu2006}, or static general TBoxes in the form of integrity constraints \cite{Giacomo2007}. The main reasons for these restrictions were expressibility and computability of the updated ontology. But we believe there is a more fundamental problem with using belief update operators to update TBoxes because it frequently yields counterintuitive results, as illustrated here:

\begin{example}[Counterintuitive TBox update] \label{ex:tbox update}
Suppose we want to update the description logic TBox $\tbox = \set{ B \sqsubseteq A }$ and we want to update it with the new information $\tbox[U] = \set{ C \sqsubseteq B }$. In other words, we introduce a new subconcept $C$ of concept $B$. Using Winslett's update operator we obtain the updated knowledge base $\set{C \sqsubseteq B, B \sqcap \lnot C \sqsubseteq A}$. Thus, the subconcept axiom from $\tbox$ is severely weakened. Using other operators (see \cite{Herzig1999} for a survey) it may even get completely forgotten. Such a forgetful behaviour cannot be explained by the sole fact that we are recording a change that occurred in the modelled environment -- new subconcepts may arise without disturbing other relations the target concept may have.
\end{example}

Thus, appropriate ways of updating ontologies in general, and TBoxes in particular, still need to be explored and pose an important open problem on its own. In our current paper we follow the mentioned ontology update literature and focus on ABox updates, leaving the TBox static throughout the update process.

Updates were also investigated in the context of Logic Programs. Earlier approaches based on literal inertia \cite{Marek1998} proved not sufficiently expressive for dealing with rule updates, leading to the development of rule update semantics based on different intuitions, principles and constructions, when compared to their classical counterparts. For example, the introduction of the \emph{causal rejection principle} \cite{Leite1997} lead to several approaches to rule updates \cite{Alferes2000,Leite2003,Eiter2002,Alferes2005}, all of them with a strong syntactic flavour which makes them very hard to combine with belief update operators that are semantic in their nature. Other existing approaches to updates of Logic Programs  \cite{Sakama2003,Zhang2005,Delgrande2008} have different problems, such as, for example, not being immune to tautological updates. It has been shown in \cite{Eiter2002} that the above mentioned rationality postulates, set forth in the context of Classical Logic, are inappropriate for dealing with updates of Logic Programs.

In order to develop an appropriate update operator for hybrid knowledge bases, one has to somehow combine these apparently irreconcilable approaches to updates, a problem that is far away from having an appropriate solution.

In this paper, we take an important first step in addressing the updates of hybrid knowledge bases. Following the state of the art in ontology updates \cite{Liu2006,Giacomo2007}, we choose a constrained scenario -- which is, nevertheless, rich enough to encompass many practical applications of hybrid theories -- in which only the ABox is allowed to evolve, while the TBox is kept static. We add rule support to this scenario by augmenting the traditional immediate consequence operator used in logic programming with the classical update operator. The resulting framework is significantly more expressive than those of \cite{Liu2006,Giacomo2007} and allows for a seamless two-way interaction between Logic Programming rules and Description Logic axioms. The consequences of rules are also subject to update through the ABox updates, making it possible to use rules to represent default preferences or behaviour and later directly impose exceptions to those rules.

The resulting update semantics enjoys several desirable properties, namely it:

\begin{itemize}
	\item generalises the stable model semantics \cite{Gelfond1988}.

	\item generalises, under reasonable assumptions, the MKNF semantics for hybrid knowledge bases \cite{Motik2007}.

	\item generalises, under reasonable assumptions, the minimal change update operator \cite{Winslett1990}.

	\item adheres to the principle of primacy of new information \cite{Dalal1988}, so every model resulting from the update by an ABox $\mathcal{A}$ is a model of $\mathcal{A}$.

	\item is syntax-independent w.r.t. the TBox and ABox, i.e. yields the same result with equivalent TBoxes and when updating by equivalent ABoxes.
\end{itemize}

To the best of our knowledge, this is the first proposal of an update semantics for hybrid knowledge bases in a single framework. This semantics not only provides an appropriate solution to the constrained scenario we chose, but it unveils a set of important issues, opening the door for interesting future research endeavours.

The remainder of this paper is structured as follows: In Sect. \ref{sect:preliminaries} we introduce the notions needed throughout the rest of the paper, and discuss some of the choices we made. Section \ref{sect:hybrid operator} contains the definition of our operator while in Sect. \ref{sect:properties} we examine its properties. In Sect. \ref{sect:discussion} we conclude and sketch some directions for future work.

\section{Preliminaries} \label{sect:preliminaries}

In this section we present the necessary preliminaries that we need to define the hybrid update operator, and discuss some of the choices we made. As the basis for the formal part of our investigation, we choose the same notation and notions as those used for Hybrid MKNF Knowledge Bases \cite{Motik2007}. This makes it possible to treat first-order formulae and nonmonotonic rules in a unified manner and also compare our semantics to the one of Hybrid MKNF more easily.

\subsection{MKNF}

The logic of minimal knowledge and negation as failure (MKNF) is an extension of first-order logic with two modal operators: $\mk[]$ and $\mnot[]$. In the following, we follow the presentation of syntax and semantics of this logic as given in \cite{Motik2007}. We use a function-free first-order syntax extended by the mentioned modal operators in a natural way. Similarly as in \cite{Motik2007}, we consider only Herbrand interpretations in our semantics.

We begin with the definition of syntax of MKNF formulas. First we need to introduce the language of MKNF:

\begin{definition}[MKNF Language]
An \emph{MKNF language} contains
\begin{enumerate*}
	\item \emph{logical connectives} $\lnot$ and $\land$;
	\item the \emph{quantifier} $\exists$;
	\item \emph{modal operators} $\mk$ and $\mnot$;
	\item \emph{punctuation symbols} ``$($'', ``$)$'' and ``$,$'';
	\item a countably infinite set of \emph{variables} $\lvar = \set{x, X, y, Y, \dotsc}$;
	\item a set of \emph{constant symbols} $\lcon = \set{c, d, \dotsc}$ and
	\item a set of \emph{predicate symbols} $\lpre = \set{P, Q, \dotsc}$, each with an associated natural number that we called its \emph{arity}.
\end{enumerate*}
Each MKNF language is determined by specifying the set of constant symbols $\lcon$ and the set of predicate symbols $\lpre$. Such a language is denoted by $\lang_{\mathrm{MKNF}}(\lcon, \lpre)$. The language is always assumed to contain at least one predicate symbol and at least one constant symbol.
\end{definition}

From now onwards, we assume that the MKNF language $\lang = \lang_{\mathrm{MKNF}}(\lcon, \lpre)$ is given and use it implicitly in the text below. Almost all the defined notions are with respect to this language but we do not stress this fact in the definitions. So instead of defining an ``MKNF formula of $\lang$'', we simply define an ``MKNF formula'', leaving out the words ``of $\lang$''. Similarly, instead of defining an ``MKNF structure over $\lang$'', we simply define an ``MKNF structure'', leaving out the words ``over $\lang$''. Other definitions follow this pattern as well.

Furthermore, while in the definitions the notions are defined with their full names (e.g. ``MKNF language'', ``MKNF formula'', \dots), further in the text we occasionally drop the word ``MKNF''. We believe these simplifications do not cause any confusion while significantly improving the readability of the text.

We continue with the definition of an MKNF formula:

\begin{definition}[MKNF Formula]
A \emph{term} is a variable or a constant. A \emph{first-order atom} is every expression of the form
\[
P(t_1, t_2, \dotsc, t_n)
\]
where $P$ is a predicate symbol of arity $n$ and each $t_i$ is a term.

The set of \emph{MKNF formulas} is the smallest set satisfying the following conditions:
\begin{enumerate}
	\item Every first-order atom is an MKNF formula.
	\item If $\phi, \psi$ are MKNF formulas and $x$ is a variable, then $\lnot \phi$, $(\phi \land \psi)$, $(\exists x : \phi)$, $\mk \phi$ and $\mnot \phi$ are also MKNF formulas.
\end{enumerate}
Where it doesn't cause confusion, the parenthesis are removed for the sake of readability. Furthermore, $(\phi \lor \psi)$, $(\phi \lthen \psi)$, $(\phi \lif \psi)$, $(\phi \lequiv \psi)$, $\ltrue$, $\lfalse$ and $(\forall x : \phi)$ are used as shortcuts for $\lnot (\lnot \phi \land \lnot \psi)$, $(\lnot \phi \lor \psi)$, $(\phi \lor \lnot \psi)$, $(\phi \lthen \psi) \land (\phi \lif \psi)$, $(p \lor \lnot p)$, $(p \land \lnot p)$ and $\lnot (\exists x : \lnot \phi)$, respectively, where $p$ is a fixed ground first-order atom from the language.\footnote{As stated in above, we assume that at least one predicate symbol and at least one constant symbol exist in the language, from which at least one ground first-order atom can be formed.}

An MKNF formula of the form $\mk \phi$ is called a \emph{modal $\mk[]$-atom}, and a formula of the form $\mnot \phi$ is called a \emph{modal $\mnot[]$-atom}; collectively, modal $\mk[]$- and $\mnot[]$-atoms are called \emph{modal atoms}. An MKNF formula $\phi$ is a \emph{sentence} if it has no free variable occurences; $\phi$ is \emph{open} if all its variable occurences are free; $\phi$ is \emph{ground} if it does not contain variables; $\phi$ is \emph{positive} if it does not contain occurrences of $\mnot[]$; $\phi$ is \emph{first-order} or \emph{objective} if it does not contain modal operators. By $\phi[t_1/x_1, t_2/x_2, \dotsc, t_n/x_n]$ we denote the formula obtained by simultaneously replacing in $\phi$ all free occurences of the variable $x_i$ by the term $t_i$ for every $i \in \set{1, 2, \dotsc, n}$.

A set of MKNF sentences is an \emph{MKNF theory}. An MKNF theory has property $X$ if all its members do (for instance, an MKNF theory is \emph{first-order} if all sentences inside it are first-order).
\end{definition}

Now we can define the semantics of MKNF formulas. We use Herbrand interpretations, assuming that apart from the constants from $\lcon$ occurring in the formulas, the signature contains a coutably infinite supply of constants not occurring in the formulas. The Herbrand Universe of such a signature is denoted by $\Delta$ and has the property $\lcon \subseteq \Delta$. If not stated otherwise, we assume that one fixed Herbrand Universe $\Delta$ with these properties is used as the universe for all interpretations.

\begin{definition}[First-Order Interpretation and Model]
A \emph{first-order interpretation} $I$ is a relational structure that contains for every predicate symbol $P \in \lpre$ of arity $n$ a relation $P^I \subseteq \Delta^n$. The set of all first-order interpretations is denoted by $\foint$.

Each first-order interpretation determines a unique truth assignment to all first-order sentences. The satisfiability of a first-order sentence $\phi$ in $I$ is defined inductively as follows:
\renewcommand{\labelenumi}{\arabic{enumi}$^\circ$}
\begin{enumerate}
	\item If $\phi$ is a ground first-order atom $P(c_1, c_2, \dotsc, c_n)$, then $\phi$ is true in $I$ if and only if $(c_1, c_2, \dotsc, c_n) \in P^I$;
	\item If $\phi$ is a first-order formula of the form $\lnot \psi$, then $\phi$ is true in $I$ if and only if $\psi$ is not true in $I$;
	\item If $\phi$ is a first-order formula of the form $\phi_1 \land \phi_2$, then $\phi$ is true in $I$ if and only if $\phi_1$ is true in $I$ and $\phi_2$ is true in $I$;
	\item If $\phi$ is a first-order formula of the form $(\exists x : \psi)$, then $\phi$ is true in $I$ if and only if $\psi[c/x]$ is true in $I$ for some constant $c \in \Delta$.
\end{enumerate}
The fact that $\phi$ is true in $I$ is denoted by $I \ent \phi$. A formula $\phi$ is false in $I$ if and only if it is not true in $I$, denoted by $I \nent \phi$. For a first-order theory $\thr$ we say that $\thr$ is true in $I$, denoted by $I \ent \thr$, if $I \ent \phi$ for each $\phi \in \thr$. Otherwise, $\thr$ is false in $I$, denoted by $I \nent \thr$.

If $I \ent \phi$, then we say that $I$ is a \emph{model of $\phi$}. Similarly, if $I \ent \thr$, then $I$ is a \emph{model of $\thr$}. The set of all models of $\phi$ is denoted by $\smod(\phi)$. The set of all models of $\thr$ is denoted by $\smod(\thr)$.
\end{definition}

The satisfiability of MKNF formulas is defined with respect to MKNF structures.

\begin{definition}[MKNF Structure]
An \emph{MKNF structure} is a triple $\mstr$ where $I$ is a first-order interpretation and $M, N$ are sets of first-order interpretations.\footnote{In difference to \cite{Motik2007}, we allow for empty $M, N$ in this definition as later on it will be useful to have satisfiability defined even for this marginal case. However, the empty set is still not considered an MKNF interpretation as can be seen further in Definition \ref{def:mknf:models}}
\end{definition}

Every MKNF structure has three components. The first is a first-order interpretation used to interpret the objective parts of a formula. The second and third are sets of first-order interpretations used to interpret the parts of a formula under the $\mk[]$ and $\mnot[]$ modality, respectively.

\begin{definition}[MKNF Satisfiability] \label{def:mknf:sat}
Let $\mstr$ be an MKNF structure. The satisfiability of an MKNF sentence $\phi$ in $\mstr$ is defined inductively as follows:
\renewcommand{\labelenumi}{\arabic{enumi}$^\circ$}
\begin{enumerate}
	\item If $\phi$ is a ground first-order atom $P(c_1, c_2, \dotsc, c_n)$, then $\phi$ is true in $\mstr$ if and only if $(c_1, c_2, \dotsc, c_n) \in P^I$;
	\item If $\phi$ is a first-order formula of the form $\lnot \psi$, then $\phi$ is true in $\mstr$ if and only if $\psi$ is not true in $\mstr$;
	\item If $\phi$ is a first-order formula of the form $\phi_1 \land \phi_2$, then $\phi$ is true in $\mstr$ if and only if $\phi_1$ is true in $\mstr$ and $\phi_2$ is true in $\mstr$;
	\item If $\phi$ is a first-order formula of the form $(\exists x : \psi)$, then $\phi$ is true in $\mstr$ if and only if $\psi[c/x]$ is true in $\mstr$ for some constant $c \in \Delta$;
	\item If $\phi$ is a formula of the form $\mk \psi$, then $\phi$ is true in $\mstr$ if and only if $\psi$ is true in $\mstr[J, M, N]$ for each $J \in M$;
	\item If $\phi$ is a formula of the form $\mnot \psi$, then $\phi$ is true in $\mstr$ if and only if $\psi$ is not true in $\mstr[J, M, N]$ for some $J \in N$.
\end{enumerate}
The fact that $\phi$ is true in $\mstr$ is denoted by $\mstr \ent \phi$. A formula $\phi$ is false in $\mstr$ if and only if it is not true in $\mstr$, denoted by $\mstr \nent \phi$.
\end{definition}

Now we are ready to introduce the notions of MKNF interpretation and model.

\begin{definition}[MKNF Interpretation and Model] \label{def:mknf:models}
An \emph{MKNF interpretation} $M$ is a non-empty set of first-order interpretations. By $\mint = 2^\foint$ we denote the set of all MKNF interpretations together with the empty set.

Let $\phi$ be an MKNF sentence, $\thr$ an MKNF theory and $M \in \mint$. We say $\phi$ is true in $M$, denoted by $M \ent \phi$, if $\mstr[I, M, M] \ent \phi$ for each $I \in M$.\footnote{Notice that if $M$ is empty, this condition is vacuously satisfied for any sentence $\phi$, so any sentence is true in $\emptyset$.} Otherwise $\phi$ is false in $M$, denoted by $M \nent \phi$. $\thr$ is true in $M$, denoted by $M \ent \thr$, if $M \ent \phi$ for each $\phi \in \thr$. Otherwise, $\thr$ is false in $M$, denoted by $M \nent \thr$.

If $M \in \mint$ is non-empty\footnote{As seen above, every formula is true in $\emptyset$, so $\emptyset$ is not considered an MKNF interpretation and for the same reason it is never given the status of a model.}, then $M$ is
\begin{itemize}
	\item an \emph{S5 model of $\phi$} if $M \ent \phi$;
	\item an \emph{S5 model of $\thr$} if $M \ent \thr$;
	\item an \emph{MKNF model of $\phi$} if $M$ is an S5 model of $\phi$ and for every MKNF interpretation $M' \supsetneq M$ there is some $I' \in M'$ such that $\mstr[I', M', M] \nent \phi$;
	\item an \emph{MKNF model of $\thr$} if $M$ is an S5 model of $\thr$ and for every MKNF interpretation $M' \supsetneq M$ there is some $I' \in M'$ and some $\phi \in \thr$ such that $\mstr[I', M', M] \nent \phi$.
\end{itemize}

If there exists the greatest S5 model of $\phi$, then it is denoted by $\smod(\phi)$. If $\phi$ has no S5 model, then $\smod(\phi)$ denotes the empty set. For the rest of MKNF sentences, $\smod(\cdot)$ stays undefined. If there exists the greatest S5 model of $\thr$, then it is denoted by $\smod(\thr)$. If $\thr$ has no S5 model, then $\smod(\thr)$ denotes the empty set. For the rest of MKNF theories, $\smod(\cdot)$ stays undefined.
\end{definition}

\subsection{Description Logics}

Description Logics (DLs) \cite{Baader2003} are (mostly) decidable fragments of first-order logic that are frequently used for knowledge representation in practical applications. In the following we assume that some Description Logic is used to describe an ontology. We do not choose any specific Description Logic, we only assume that the ontology expressed in it is composed of two distinguishable parts: a TBox with concept and role definitions using the constructs of the underlying description logic, and an ABox with individual assertions, i.e. assertions of the form $C(a)$ and $R(a, b)$ where $a, b$ are constants, $C$ is a concept expression and $R$ is a role expression of the un	derlying description logic. This distinction is important to us as we treat the two types of knowledge in different ways -- the TBox is considered static while the ABox is allowed to evolve. As was noted in the introduction, our main reason for this is that we believe existing update operators to be unsuitable for updating concept definitions contained in the TBox. We also assume that the axioms of the underlying DL can be translated into first-order logic and for the sake of simplicity we assume that the TBox and ABox already contain these translations instead of the syntactic constructs of the underlying DL.

\subsection{Hybrid MKNF Knowledge Bases}

We make use of the general MKNF framework to give a semantics to hybrid knowledge bases composed of an ontology and a normal logic program. The following definition introduces the notion of a rule as we use it in the following:

\begin{definition}[Rule]
A \emph{rule} is any open MKNF formula of the form
\begin{equation} \label{eq:mknf_programs:mknf_rule}
\mk p \lif \mk q_1 \land \mk q_2 \land \dotsb \land \mk q_k \land \mnot s_1 \land \mnot s_2 \land \dotsb \land \mnot s_l
\end{equation}
where $k, l$ are non-negative integers and $p, q_i, s_j$ are first-order atoms for any $i \in \set{1, 2, \dotsc, k}, j \in \set{1, 2, \dotsc, l}$. Given a rule $r$ of the form \eqref{eq:mknf_programs:mknf_rule}, the following notation is also defined:
\begin{align*}
H(r) &= \mk p \enspace, \\
H^*(r) &= p \enspace, \\
B^+(r) &= \Set{\mk q_1, \mk q_2, \dotsc, \mk q_k} \enspace, \\
B^-(r) &= \Set{\mnot s_1, \mnot s_2, \dotsc, \mnot s_l} \enspace, \\
B(r) &= B^+(r) \cup B^-(r) \enspace.
\end{align*}
$H(r)$ is dubbed the \emph{head of $r$}, $H^*(r)$ the \emph{first-order head of $r$}, $B^+(r)$ the \emph{positive body of $r$}, $B^-(r)$ the \emph{negative body of $r$} and $B(r)$ the \emph{body of $r$}. A rule $r$ is called \emph{definite} if its negative body is empty. A rule $r$ is called a \emph{fact} if its body is empty.

A \emph{program} is a set of rules. A \emph{definite program} is a set of definite rules.
\end{definition}

As was shown in \cite{Lifschitz1991}, the MKNF semantics generalises the stable model semantics for logic programs. In particular, every logic programming rule of the form
\[
p \lpif q_1, q_2, \dotsc, q_k, \lpnot s_1, \lpnot s_2, \dotsc, \lpnot s_l.
\]
can be translated into the MKNF formula \eqref{eq:mknf_programs:mknf_rule} and the stable models of sets of such rules (i.e. of normal logic programs) directly correspond to MKNF models of the set of translated rules.

We are now ready to define a hybrid knowledge base and its semantics.

\begin{definition}[Hybrid knowledge base]
Let $\ont$ be an ontology and $\prog$ a program. The pair $\kb = \an{\ont, \prog}$ is then called a \emph{hybrid knowledge base}. We say $\kb$ is \emph{definite} if $\prog$ is definite and we say $\kb$ is \emph{$\prog$-ground} if $\prog$ is ground.
\end{definition}

The semantics of hybrid knowledge bases is given in terms of a translation $\pi$ into a set of MKNF formulas which is defined as follows:

\begin{definition}
For an ontology $\ont$, a rule $r$ with the vector of free variables $\mathbf{x}$, a program $\prog$ and the hybrid knowledge base $\kb = \an{\ont, \prog}$, we define:
\begin{align*}
\pi(\ont) &= \Set{ \mk \phi | \phi \in \ont } \enspace, \\
\pi(r) &= (\forall \mathbf{x} : r) \enspace, \\
\pi(\prog) &= \Set{ \pi(r) | r \in \prog } \enspace, \\
\pi(\kb) &= \pi(\ont) \cup \pi(\prog) \enspace.
\end{align*}
We say an MKNF interpretation $M$ is an \emph{S5 model of $\kb$} if $M$ is an S5 model of $\pi(\kb)$. We say $M$ is an \emph{MKNF model of $\kb$} if $M$ is an MKNF model of $\pi(\kb)$.
\end{definition}

In this paper, we are not concerned with decidability of reasoning, so we refrain from introducing a safety condition on our rules as was done in \cite{Motik2007}.

\subsection{Classical Updates}
As a basis for our update operator, we adopt an update semantics called the \emph{minimal change update semantics} (sometimes also called the \emph{possible models approach} (PMA)) as defined in \cite{Winslett1990} for updating first-order theories. There are a number of reasons for this choice. First, it satisfies all of Katsuno and Mendelzon's update postulates \cite{Katsuno1991}. This means, for instance, that unlike some other update semantics, such as the standard semantics \cite{Winslett1990}, it is not sensitive to syntax of the original theory or of the update. Second, it is based on an intuitive idea, treating each classical model of the original theory as a possible world and modifying it as little as possible in order to become consistent with the new information. This idea has its roots in reasoning about action \cite{Winslett1988} and updates of relational theories \cite{Winslett1990}. Third, the operator has already been successfully used to deal with ABox updates \cite{Liu2006,Giacomo2007}.

This semantics uses a notion of closeness of first-order interpretations w.r.t. a fixed first-order interpretation $I$. This notion is based on the set of ground first-order atoms that are interpreted differently than in $I$.

\begin{definition}[Interpretation distance]
Let $P$ be a predicate symbol and $I, J$ be first-order interpretations. The \emph{difference in the interpretation of $P$ between $I$ and $J$}, written $\mathit{diff}(P, I, J)$, is a relation containing the set of tuples $(P^I \setminus P^J) \cup (P^J \setminus P^I)$.

Given first-order interpretations $I, J, J'$, we say that $J$ is at least as close to $I$ as $J'$, denoted by $J \leq_I J'$, if for every predicate symbol $P$ it holds that $\mathit{diff}(P, I, J)$ is a subset of $\mathit{diff}(P, I, J')$. We also say that \emph{$J$ is closer to $I$ than $J'$}, denoted by $J <_I J'$, if $J \leq_I J'$ and $J' \nleq_I J$.
\end{definition}

We now give a definition of the minimal change update semantics but in difference to \cite{Winslett1990}, we use a specific vocabulary which is closer to the setting of this paper. In particular, we define the semantics of updating an initial theory $\thr$ by an ABox $\abox$ in the context of the TBox $\tbox$. The TBox is treated as static integrity constraints for the whole update process. The minimal change update semantics chooses those models of $\tbox \cup \abox$ that are the closest w.r.t. the relation $\leq_I$ to some model $I$ of $\tbox \cup \thr$. Formally:

\begin{definition}[Winslett's minimal change update semantics] \label{def:classical updates:minimal change}
Let $\thr$ be a first-order theory, $\tbox$ a TBox, $\abox$ an ABox, $I$ a first-order interpretation and $M$ a set of first-order interpretations. We define:
\begin{align*}
\mathsf{incorporate}^{\tbox}(\abox, I) &= \Set{ J \in \smod(\tbox \cup \abox) | (\nexists J' \in \smod(\tbox \cup \abox))(J' <_I J) } \enspace, \\
\mathsf{incorporate}^{\tbox}(\abox, M) &= \bigcup_{I \in M} \mathsf{incorporate}^{\tbox}(\abox, I) \enspace, \\
\smod(\thr \oplus^{\tbox} \abox) &= \mathsf{incorporate}^{\tbox}(\abox, \smod(\tbox \cup \thr)) \enspace.
\end{align*}
If $\smod(\thr \oplus^{\tbox} \abox)$ is nonempty, we call it the \emph{minimal change update model of $\thr \oplus^{\tbox} \abox$}.
\end{definition}

The previous definition can be naturally generalised to allow for sequences of ABoxes. Starting from the models of the original theory, for each ABox in the sequence we transform the set of models according to the minimal change update semantics defined above. The resulting set of models then determines the updated theory. Formally:

\begin{definition}[Update by a sequence of ABoxes] \label{def:classical updates:iterative minimal change}
Let $\thr$ be a first-order theory, $\tbox$ a TBox, $\abox = (\abox_1, \abox_2, \dotsc, \abox_n)$ a sequence of ABoxes and $M$ a set of first-order interpretations. We inductively define:
\begin{align*}
\mathsf{incorporate}^{\tbox}(\abox, M) &= \mathsf{incorporate}^{\tbox}((\abox_2, \dotsc, \abox_n), \mathsf{incorporate}^{\tbox}(\abox_1, M)) \enspace, \\
\smod(\thr \oplus^{\tbox} \abox) &= \mathsf{incorporate}^{\tbox} (\abox, \smod(\tbox \cup \thr)) \enspace.
\end{align*}
If $\smod(\thr \oplus^{\tbox} \abox)$ is nonempty, we call it the \emph{minimal change update model of $\thr \oplus^{\tbox} \abox$}.
\end{definition}

\section{Hybrid Update Operator} \label{sect:hybrid operator}

Turning to the formal part of our proposal, our aim is to propose a semantics for a program $\prog$ updated by a sequence of ABoxes $(\abox_1, \abox_2, \dotsc, \abox_n)$ in the context of a TBox $\tbox$. We assume program $\prog$ to be finite and ground, a common assumption when dealing with reasoning under the stable model semantics.

We follow a path similar to how the stable models of normal logic programs were originally defined \cite{Gelfond1988}, and start by defining how a definite program can be updated by a sequence of ABoxes, and only afterwards deal with programs containing default negation.

As with the least model of a definite logic program, our resulting model is the least fixed point of an immediate consequence operator. Our operator is in a way similar to the usual immediate consequence operator $T_{\prog}$ commonly used to draw consequences from a logic program $\prog$. The crucial difference between $T_{\prog}$ and our operator is that in the latter, the consequences are subsequently updated by the sequence of ABoxes $\abox$ using the classical update operator. Formally:

\begin{definition}[Updating immediate consequence operator $T_{\prog \oplus^{\tbox} \abox}$]
Let $\prog$ be a finite ground definite program, $\tbox$ a TBox and $\abox$ a sequence of ABoxes. We define the operator $T_{\prog \oplus^{\tbox} \abox}$ for any $M \subseteq \foint$ as follows\footnote{Recall that $M \ent B(r)$ holds if and only if $M$ is an S5 model of every modal atom in $B(r)$ (see also Def. \ref{def:mknf:models}).}:
\begin{align*}
T_{\prog \oplus^{\tbox} \abox}(M) &= \smod(\Set{ H^*(r) | r \in \prog \land M \ent B(r) } \oplus^{\tbox} \abox)
\end{align*}
\end{definition}

An important property of an immediate consequence operator is \emph{continuity} because it guarantees the existence of a least fixed point and also provides a way of computing this least fixed point (using the Kleene Fixed Point Theorem). The $T_{\prog \oplus^{\tbox} \abox}$ operator satisfies the condition of continuity:

\begin{proposition}[Continuity of $T_{\prog \oplus^{\tbox} \abox}$]
\label{prop:hybrid update operator:tpu continuous}
Let $\prog$ be a finite ground definite program, $\tbox$ a TBox and $\abox$ a sequence of ABoxes. Then $T_{\prog \oplus^{\tbox} \abox}$ is a continuous function on the complete partial order  $(\mint, \supseteq)$.
\end{proposition}
\begin{proof}
See Appendix \ref{app:hybrid update operator}, page \pageref{proof:hybrid update operator:tpu continuous}.
\end{proof}

Now we can define a \emph{minimal change dynamic stable model} of $\prog \oplus^{\tbox} \abox$, where $\prog$ is a definite program, as the least fixed point of $T_{\prog \oplus^{\tbox} \abox}$:

\begin{definition}[Minimal change dynamic stable model for definite programs]
Let $\prog$ be a finite ground definite program, $\tbox$ a TBox and $\abox$ a sequence of ABoxes. We say an MKNF interpretation $M$ is a \emph{minimal change dynamic stable model of $\prog \oplus^{\tbox} \abox$} if it is the least fixed point of $T_{\prog \oplus^{\tbox} \abox}$.
\end{definition}

Notice that for every definite program $\prog$ and each sequence of ABoxes $\abox$, $\prog \oplus^{\tbox} \abox$ has either no minimal change dynamic stable model (when the least fixed point of $T_{\prog \oplus^{\tbox} \abox}$ is empty), or exactly one minimal change dynamic stable model.

In order to deal with default negation in the bodies of rules, we use the Gelfond-Lifschitz transformation which was used to define the stable models of a normal logic program \cite{Gelfond1988}. We do this by defining the definite program $\progmod$ which is the result of performing the Gelfond-Lifschitz transformation on $\prog$ -- rules from $\prog$ with a negative body that is in conflict with $M$ are discarded, while for all the other rules, their negative bodies are discarded. Then $\progmod$ is updated by $\abox$ using the above definition for definite logic programs and if the result is identical to $M$, then $M$ is given the status of a \emph{minimal change dynamic stable model}. Hence, the resulting operator can be used to update an arbitrary normal logic program by a sequence of ABoxes.

\begin{definition}[Minimal change dynamic stable model]
Let $\prog$ be a finite ground program, $\tbox$ a TBox, $\abox$ a sequence of ABoxes and $M$ an MKNF interpretation. We say $M$ is a \emph{minimal change dynamic stable model of $\prog \oplus^{\tbox} \abox$} if $M$ is a minimal change dynamic stable model of $\progmod \oplus^{\tbox} \abox$ where
\[
\progmod = \Set{ H(r) \lif B^+(r) | r \in \prog \land M \ent B^-(r) } \enspace.
\]
\end{definition}

The minimal change dynamic stable models can be used to define a consequence relation from $\prog \oplus^{\tbox} \abox$ where $\prog$ is a finite ground program, $\tbox$ is a TBox and $\abox$ a sequence of ABoxes. We offer a definition which adopts a skeptical approach to inference, credulous and other definitions may be obtained similarly.

\begin{definition}[Consequence relation] \label{def:consequence relation}
Let  $\prog$ be a finite ground program, $\tbox$ a TBox, $\abox$ a sequence of ABoxes and $\phi$ an MKNF sentence. We say that \emph{$\prog \oplus^{\tbox} \abox$ entails $\phi$}, written $\prog \oplus^{\tbox} \abox \ent \phi$, if and only if $M \ent \phi$ for all minimal change dynamic stable models $M$ of $\prog \oplus^{\tbox} \abox$.
\end{definition}

We now demonstrate the defined update semantics on a simple example:

\begin{example} \label{ex:hybrid operator}
Consider the following TBox $\tbox$ and program $\prog$:
\begin{align}
\tbox: \quad
& A \equiv B \sqcup C \label{eq:hybrid operator:tbox:1} \\
& \mathit{NegA} \equiv \lnot A \label{eq:hybrid operator:tbox:2} \\
& D \equiv \lnot A \sqcap \exists P^-.A \label{eq:hybrid operator:tbox:3} \\
\prog: \quad
& \mathit{NegA}(X) \lpif \lpnot A(X). \label{eq:hybrid operator:program:1} \\
& P(X, Y) \lpif A(X), E(Y), \lpnot E(X). \label{eq:hybrid operator:program:2}
\end{align}
TBox assertions \eqref{eq:hybrid operator:tbox:1} and \eqref{eq:hybrid operator:tbox:2} together with rule \eqref{eq:hybrid operator:program:1} define the concept $A$ as a union of concepts $B$ and $C$ and they make this concept interpreted under CWA instead of OWA, i.e. whenever for some constant $c$ we cannot conclude that $A(c)$ is true, the rule \eqref{eq:hybrid operator:program:1} infers $\mathit{NegA}(c)$ and by \eqref{eq:hybrid operator:tbox:2} we obtain $\lnot A(c)$. Assertion \eqref{eq:hybrid operator:tbox:3} defines concept $D$ as those members $d$ of $\lnot A$ for which there exists some $c$ from $A$ with $P(c, d)$. Rule \eqref{eq:hybrid operator:program:2} infers the relation $P(c, d)$ whenever $c$ is in $A$ but not in $E$ and $d$ is in $E$.

Given the initial definitions, an update by $\abox_1 = \set{A(c)}$ now yields\footnote{In the example we assume that the rules are grounded using all constants explicitly mentioned in the knowledge base. In this case there are only two: $c$ and $d$.}
\[
\prog \oplus^{\tbox} \abox_1 \ent \set{A(c), \lnot A(d)} \enspace.
\]
A further update by $\abox_2 = \set{\lnot B(c)}$ introduces a possibility of $A(c)$ not being true in case $B(c)$ was true before and $C(c)$ was false. Since $A$ is interpreted under the closed world assumption, we can now conclude that $A(c)$ is false:
\[
\prog \oplus^{\tbox} (\abox_1, \abox_2) \ent \set{\lnot A(c), \lnot B(c), \lnot A(d)}
\]
Consider now the update $\abox_3 = \set{C(c) \land E(d)}$. Given \eqref{eq:hybrid operator:tbox:1}, this reinstates $A(c)$. Furthermore, rule \eqref{eq:hybrid operator:program:2} can now infer $P(c, d)$ and by \eqref{eq:hybrid operator:tbox:2} we obtain $D(d)$:
\[
\prog \oplus^{\tbox} (\abox_1, \abox_2, \abox_3) \ent \set{A(c), \lnot B(c), C(c), \lnot A(d), E(d), P(c, d), D(d)}
\]
In the next update $\abox_4 = \set{E(c)}$ we block the body of rule \eqref{eq:hybrid operator:program:2}, which also prevents $D(d)$ from being inferred:
\[
\prog \oplus^{\tbox} (\abox_1, \abox_2, \abox_3, \abox_4) \ent \set{A(c), \lnot B(c), C(c), \lnot A(d), E(d), E(c)}
\]
The last update\footnote{Updating ABoxes could, of course, be more complex since arbitrary concept expressions may be used (e.g. $(\exists P.C)(c)$). Here, due to limited space, we keep the example very simple.} $\abox_5 = \set{\lnot E(c) \land \lnot P(c, d)}$ illustrates how the conclusion of a rule may be overridden through the ABox updates -- though the body of rule \eqref{eq:hybrid operator:program:2} is true, its head does not become true since it is in direct conflict with $\abox_5$:
\[
\prog \oplus^{\tbox} (\abox_1, \abox_2, \abox_3, \abox_4, \abox_5) \ent \set{A(c), \lnot B(c), C(c), \lnot A(d), E(d), \lnot E(c), \lnot P(c, d)}
\]
\end{example}

\section{Properties and Relations} \label{sect:properties}

In this section we investigate a number of formal properties of the defined operator. The first property guarantees that every minimal change dynamic stable model of $\prog \oplus^{\tbox} \abox$ is a model of $\abox$. This is known as the \emph{principle of primacy of new information} \cite{Dalal1988}.

\begin{proposition}[Primacy of new information]
\label{prop:hybrid update operator:KM:1}
Let $\prog$ be a finite ground program, $\tbox$ a TBox, $\abox$ an ABox and $M$ a minimal change dynamic stable model of $\prog \oplus^{\tbox} \abox$. Then $M \ent \abox$.
\end{proposition}
\begin{proof}
See Appendix \ref{app:hybrid update operator}, page \pageref{proof:hybrid update operator:KM:1}.
\end{proof}

The second property guarantees that our operator is syntax-independent w.r.t. the TBox and the updating ABox. This is a desirable property as it shows that providing equivalent TBoxes and updating by equivalent ABoxes always produces the same result. It is inherited from the classical minimal change update operator.

\begin{proposition}[Syntax independence]
\label{prop:hybrid update operator:KM:4}
Let $\prog$ be a finite ground program, $\tbox, \tbox'$ be TBoxes such that $\smod(\tbox) = \smod(\tbox')$, $\abox, \abox'$ be ABoxes such that $\smod(\abox) = \smod(\abox')$ and $M$ be an MKNF interpretation. Then $M$ is a minimal change dynamic stable model of $\prog \oplus^{\tbox} \abox$ if and only if $M$ is a minimal change dynamic stable model of $\prog \oplus^{\tbox'} \abox'$.
\end{proposition}
\begin{proof}
See Appendix \ref{app:hybrid update operator}, page \pageref{proof:hybrid update operator:KM:4}.
\end{proof}

The following proposition relates the hybrid update operator to the static MKNF semantics of hybrid knowledge bases. It gives sufficient conditions for the static and dynamic semantics to coincide. In particular, the sufficient condition requires that for any set of consequences $S$ of program $\prog$ in the context of a model $M$, updating $S$ by $\abox$ in the context of $\tbox$ has the same effect as making an intersection of the models of $S$ with the models of $\abox$ and $\tbox$.

\begin{proposition}[Relation to Hybrid MKNF] \label{prop:hybrid update operator:mknf}
Let $\prog$ be a finite ground program, $\ont = \tbox \cup \abox$ an ontology with TBox $\tbox$ and ABox $\abox$ and $M$ an MKNF interpretation such that for every subset $S$ of the set $\set{ H^*(r) | r \in \prog \land M \ent B(r) }$ the following condition is satisfied:
\[
\smod( S \oplus^{\tbox} \abox ) = \smod( S \cup \ont ) \enspace.
\]
Then $M$ is an MKNF model of $\an{\ont, \prog}$ if and only if $M$ is a minimal change dynamic stable model of $\prog \oplus^{\tbox} \abox$.
\end{proposition}
\begin{proof}
See Appendix \ref{app:hybrid update operator}, page \pageref{proof:hybrid update operator:mknf}.
\end{proof}

The precondition of this proposition is satisfied, for example, when predicates appearing in heads of $\prog$ do not appear in the ontology $\ont$. An important subcase of this is when $\ont$ is empty because then the proposition implies that the minimal change dynamic stable models of $\prog \oplus^{\emptyset} \emptyset$ are exactly the MKNF models of $\prog$. Since the MKNF semantics generalises the stable model semantics \cite{Lifschitz1991}, the minimal change dynamic stable models of $\prog \oplus^{\emptyset} \emptyset$ also coincide with the stable models of $\prog$. In other words, our operator properly generalises stable models.

\begin{corollary}[Generalisation of stable models] \label{cor:hybrid update operator:mknf:sm}
Let $\prog$ be a finite ground program. Then $M$ is a stable model of $\prog$ if and only if $M$ is a minimal change dynamic stable model of $\prog \oplus^{\emptyset} \emptyset$.
\end{corollary}
\begin{proof}
See Appendix \ref{app:hybrid update operator}, page \pageref{proof:hybrid update operator:mknf:sm}.
\end{proof}

Turning to relations with the minimal change update operator, we show that updating any logic program that can be equivalently translated into first-order logic has the same effect as updating the translated first-order theory using the minimal change update operator. Hence, our update operator generalises the classical minimal change update operator.

\begin{proposition}[Generalisation of the minimal change update operator] \label{prop:hybrid update operator:generalisation of classical updates}
Let $\prog$ be a finite ground program containing only facts, $\tbox$ a TBox, $\abox$ a sequence of ABoxes and $M$ an MKNF interpretation. Then $M$ is a minimal change dynamic stable model of $\prog \oplus^{\tbox} \abox$ if and only if $M$ is a minimal change update model of $\thr_{\prog} \oplus^{\tbox} \abox$ where $\thr_{\prog} = \set{ p | \mk p \in \prog }$.
\end{proposition}
\begin{proof}
See Appendix \ref{app:hybrid update operator}, page \pageref{proof:hybrid update operator:generalisation of classical updates}.
\end{proof}

Another property that our operator inherits from the classical minimal change update operator is that empty ABoxes in the updating sequence do not influence the resulting models. Similarly, updating an empty program simply yields the set of all first-order models of $\tbox \cup \abox$. These last two properties ensure that empty program and updates cannot influence the resulting models under our update operator\footnote{Perhaps surprisingly, as shown in \cite{Leite2003}, these two properties are violated by many update operators in the context of Logic Programming.}.

\begin{proposition}[Indifference to empty updates] \label{prop:hybrid update operator:empty theory}
Let $\prog$ be a finite ground program, $\tbox$ be a TBox and $\abox = (\abox_1, \abox_2, \dotsc, \abox_n)$ a sequence of ABoxes (where $n \geq 1$). Let
\[
\abox' = (\abox_1, \abox_2, \dotsc, \abox_{i-1}, \abox_i, \emptyset, \abox_{i+1}, \dotsc, \abox_n)
\]
for some $i \in \set{0, 1, 2, \dotsc, n}$. Then an MKNF interpretation $M$ is a minimal change dynamic stable model of $\prog \oplus^{\tbox} \abox$ if and only if $M$ is a minimal change dynamic stable model of $\prog \oplus^{\tbox} \abox'$.
\end{proposition}
\begin{proof}
See Appendix \ref{app:hybrid update operator}, page \pageref{proof:hybrid update operator:mknf:sm}.
\end{proof}
\begin{proof}
See Appendix \ref{app:hybrid update operator}, page \pageref{proof:hybrid update operator:empty program}.
\end{proof}

\begin{proposition}[Updating an empty program] \label{prop:hybrid update operator:empty program}
Let $\tbox$ be a TBox, $\abox$ an ABox and $M$ an MKNF interpretation. Then $M$ is a minimal change dynamic stable model of $\emptyset \oplus^{\tbox} \abox$ if and only if $M = \smod(\tbox \cup \abox)$.
\end{proposition}

\subsubsection*{Relation to Katsuno and Mendelzon's postulates}
In the following we briefly discuss the relation of our operator to Katsuno and Mendelzon's postulates for updates of propositional knowledge bases formulated in \cite{Katsuno1991}. Each propositional knowledge base over a finite language can be represented by a single propositional formula and the result of the update can also be represented as a propositional formula. The eight desirable properties of an update operator $\diamond$ are as follows:

\textbf{KM~1}: $\phi \diamond \psi$ implies $\psi$.

\textbf{KM~2}: If $\phi$ implies $\psi$, then $\phi \diamond \psi$ is equivalent to $\phi$.

\textbf{KM~3}: If both $\phi$ and $\psi$ are satisfiable, then $\phi \diamond \psi$ is satisfiable.

\textbf{KM~4}: If $\phi_1$ is equivalent to $\phi_2$ and $\psi_1$ is equivalent to $\psi_2$, then $\phi_1 \diamond \psi_1$ is equivalent to $\phi_2 \diamond \psi_2$.

\textbf{KM~5}: $(\phi \diamond \psi) \land \chi$ implies $\phi \diamond (\psi \land \chi)$.

\textbf{KM~6}: If $\phi \diamond \psi_1$ implies $\psi_2$ and $\phi \diamond \psi_2$ implies $\psi_1$, then $\phi \diamond \psi_1$ is equivalent to $\phi \diamond \psi_2$.

\textbf{KM~7}: If for each atom $p$ either $\phi$ implies $p$ or $\phi$ implies $\lnot p$, then $(\phi \diamond \psi_1) \land (\phi \diamond \psi_2)$ implies $\phi \diamond (\psi_1 \lor \psi_2)$.

\textbf{KM~8}: $(\phi_1 \lor \phi_2) \diamond \psi$ is equivalent to $(\phi_1 \diamond \psi) \lor (\phi_2 \diamond \psi)$.

In order to examine these postulates in our setting, we restrict our attention to a finite propositional language. In order to interpret the postulates in our setting, we need to define the semantics of a number of notions used in them. Let $\prog, \prog_1, \prog_2$ be programs, $\tbox$ a TBox and $\alpha, \alpha_1, \alpha_2$ be propositional formulae representing ABox updates. We need to discuss and define, at least:
\begin{enumerate}
	\item When does $\prog \oplus^{\tbox} \alpha_1$ imply $\alpha_2$? (used in KM~1 and KM~6)
	\item When does $\prog$ imply $\alpha$? (used in KM~2 and KM~7)
	\item When is $\prog_1 \oplus^{\tbox} \alpha$ equivalent to $\prog_2$? (used in KM~2)
	\item When is $\prog$ satisfiable? (used in KM~3)
	\item When is $\prog \oplus^{\tbox} \alpha$ satisfiable? (used in KM~3)
	\item When is $\prog_1$ equivalent to $\prog_2$? (used in KM~4)
	\item When is $\prog_1 \oplus^{\tbox} \alpha_1$ equivalent to $\prog_2 \oplus^{\tbox} \alpha_2$? (used in KM~4 and KM~6)
	\item What is the semantics of $(\prog \oplus^{\tbox} \alpha_1) \land \alpha_2$? (used in KM~5)
	\item What is the semantics of $(\prog \oplus^{\tbox} \alpha_1) \land (\prog \oplus^{\tbox} \alpha_2)$? (used in KM~7)
	\item What is the semantics of $\prog_1 \lor \prog_2$? (used in KM~8)
\end{enumerate}
Most of these questions can be answered in multiple different ways while some of them are hard to provide answers to at all. In the following, we suggest ways of answering most of these questions and then analyse whether our operator satisfies the corresponding postulates.

Question 1. can be answered using the consequence relation from Def. \ref{def:consequence relation}. A similar consequence relation can be defined using stable models to answer question 2. A simple answer to question 3. is to say that $\prog_1 \oplus^{\tbox} \alpha$ is equivalent to $\prog_2$ if the set of minimal change dynamic stable models of $\prog_1 \oplus^{\tbox} \alpha$ is equal to the set of stable models of $\prog_2$. Regarding questions 4. and 5., we can say that $\prog$ is satisfiable if it has at least one stable model and $\prog \oplus^{\tbox} \alpha$ is satisfiable if it has at least one minimal change dynamic stable model. Question 6. can be answered similarly as question 3. by comparing the sets of minimal change dynamic stable models of $\prog \oplus^{\tbox} \alpha_1$ and $\prog \oplus^{\tbox} \alpha_2$. Finally, question 7. can be answered by comparing the sets of stable models of $\prog_1$ and $\prog_2$ or by using strong equivalence \cite{Lifschitz2001}. Providing reasonable answers to the remaining questions requires more investigation, so, for now, we do not further examine postulates KM~5, KM~7 and KM~8.

Turning to the rest of the postulates, we note that our operator adheres to KM~1, which was proved in Proposition \ref{prop:hybrid update operator:KM:1}. The same is not the case with postulate KM~2, as shown by the following counterexample. Consider the program
\begin{equation} \label{eq:cumulativity}
\begin{alignedat}{4}
\prog: \quad
&& p &\lpif \lpnot q. &  \qquad r &\lpif q, \lpnot r. \\
&& q &\lpif \lpnot p. & r &\lpif p.
\end{alignedat}
\end{equation}
and an update $\alpha = r$. The only stable model of $\prog$ is the maximal S5 model $M$ of $\set{p, r}$. Clearly, $M \ent \alpha$. But $\prog \oplus^{\tbox} \alpha$ has another minimal change dynamic stable $M'$, which is the maximal S5 model of $\set{q, r}$ and so is not equivalent to $\prog$.

In fact, this behaviour is inherited from the stable semantics for logic programs which does not satisfy the very similar property of \emph{cumulativity} \cite{Makinson1988,Dix1995}. Hence, it is expectable that KM~2 is never satisfied by any update semantics that properly generalises the stable model semantics.

A similar situation arises with postulate KM~3 because the stable model semantics allows to express integrity constraints, and these may easily be broken by an update. For example, the program $\prog = \set{p \lpif q, \lpnot p.}$, updated by $\alpha = q$, of which both are satisfiable, does not allow for any minimal change dynamic stable model. It is not clear how an integrity constraint should be updated because, once it is a part of the knowledge base, which is assumed to be a correct representation of the world, it should not be violated, and no new information should have the power to override it. Or should it? That is another open research question worth investigating.

Postulate KM~4 is partially formulated in Proposition \ref{prop:hybrid update operator:KM:4}, which shows that updating by equivalent ABoxes produces the same result. The other half amounts to proving that updating equivalent logic programs by the same ABox also produces equivalent results. For the two notions of program equivalence that we proposed above, this property does not hold. As a counterexample take $\prog_1 = \set{p., q.}$ and $\prog_2 = \set{p., q \lpif p.}$ which have the same answer sets and are also strongly equivalent. An update by $\alpha = \lnot p$, produces different results for $\prog_1$ and $\prog_2$, respectively, which we believe is in accord with intuitions regarding these two programs. It may be the case that for different notions of program equivalence that better suit our scenario, such as the update equivalence of logic programs proposed in \cite{Leite2003}, this property holds. Further investigation is needed to answer this question.

Finally, postulate KM~6 is also not satisfied by the operator. As a counterexample we can take the program $\prog$ defined in \eqref{eq:cumulativity}, $\alpha_1 = r$ and $\alpha_2 = p \lor q$. Then $\prog \oplus^{\tbox} \alpha_1$ has two minimal change dynamic stable models: $M_1 = \smod(\set{p, r})$ and $M_2 = \smod(\set{q, r})$. Hence, $P \oplus^{\tbox} \alpha_1 \ent \alpha_2$. Furthermore, $\prog \oplus^{\tbox} \alpha_2$ has only one minimal change dynamic stable model which is $M_1$ and consequently $\prog \oplus^{\tbox} \alpha_2 \ent \alpha_1$. However, $\prog \oplus^{\tbox} \alpha_1$ is not equivalent to $\prog \oplus^{\tbox} \alpha_2$.

\section{Conclusion and Future Work} \label{sect:discussion}

As seen, our operator properly generalises the two main ingredients that it is motivated by -- the stable model semantics of normal logic programs (Corollary \ref{cor:hybrid update operator:mknf:sm}) and the minimal change update operator (Proposition \ref{prop:hybrid update operator:generalisation of classical updates}). The failure of our operator to satisfy many of Katsuno and Mendelzon's postulates is not surprising. A wide range of classical update and revision postulates was already studied in the context of rule updates, only to find that many of them were inappropriate for characterising plausible rule update operators \cite{Eiter2002}. Furthermore, in \cite{Slota2010} we show that even under the SE model semantics, which is strictly more expressive than stable models semantics, update operators satisfying only some of the basic Katsuno and Mendelzon's postulates necessarily violate the property of support which is at the core of most logic programming semantics. The search for desirable properties of hybrid update operators is an interesting future research area.

There are also many more properties still to be examined, among them decidability as well as complexity of reasoning. Since we cannot expect the operator to perform any better than the stable model semantics and the classical update operator it is based on, its tractable approximations need to be defined and examined. The well-founded semantics for logic programs \cite{Gelder1991} and its version for hybrid MKNF knowledge bases \cite{Alferes2009} constitute crucial starting points. The recent research on ontology evolution (see \cite{Flouris2008} for a survey) can help design tractable update operators which, at the same time, offer the necessary functionality to be interesting for use in practice.

In this paper, the TBox was considered static and was treated in the same way as integrity constraints in \cite{Winslett1990}. This approach to handling integrity constraints in the context of updates has been criticized in the literature \cite{Herzig1999,Herzig2005}, as in certain cases it does not provide the expected results. However, the proposed solutions are defined only for the propositional case and a preliminary examination showed that their treatment of equivalences, such as the TBox definitions used in Example \ref{ex:hybrid operator}, is not always the expected one. Further investigation is needed to find suitable solutions to these problems in the context of ontology updates. Furthermore, in truly dynamic environments, the TBox should also be allowed to be updated. We believe that finding appropriate update operators for ontologies is still a largely open research question.

The large body of work on rule updates \cite{Leite2003,Alferes2005}, and more recently \cite{Delgrande2008}, also needs to be exploited in the attempts to define an update operator that can deal with the evolution of both rules and ontologies.

Finally, while incorporating new knowledge in a knowledge base is important, the complementary task of removing a certain piece of information is also important. Hence, hybrid erasure operators should be studied and related to hybrid update operators. The work on erasure \cite{Giacomo2007} in description logics as well as forgetting in both description logics \cite{Wang2009} and logic programs \cite{Eiter2008a} should be the starting points of this research.

To conclude, in this paper, to the best of our knowledge, we proposed the first update operator for hybrid knowledge bases. We deal with a constrained but interesting scenario in which a TBox and nonmonotonic rules represent static knowledge, policies, norms and default preferences, and the evolving ABox represents the open and dynamic environment. We illustrated the behaviour of our operator on a simple example. The operator can be used in realistic scenarios where the general notions and rules are relatively fixed, and individuals tend to change their state frequently. This is the case of many real life institutions where stakeholders change their state on a regular basis while the general rules and structures change only occasionally.

We proved a number of properties of our operator, among which its relations with the theories it was based on, such as the stable model semantics for logic programs \cite{Gelfond1988}, the MKNF semantics for hybrid knowledge bases \cite{Motik2007} and Winslett's minimal change update operator \cite{Winslett1990}.

We believe that this new area of research brings exciting new problems to solve and bridges a number of existing research areas. It will certainly provide useful results for many applications and perhaps even contribute to finding further philosophical insights into how human knowledge evolves.

\bibliographystyle{acmtrans}
\bibliography{bibliography}

\appendix 

\renewcommand{\thetheorem}{\Alph{section}.\arabic{theorem}}

\section{Kleene Fixed Point Theorem} \label{app:kleene}

Fixed points play an important role in many of the investigations in the area of logic programming. Many semantics of logic programs are defined by a fixed point equation, meaning that in order for an interpretation $M$ to be considered a ``good'' model of a logic program, it must satisfy some equation of the form $M = f(M)$ where $f$ is a mapping from interpretations to interpretations, also called an \emph{operator}. Such operators were heavily studied in Order Theory and Kleene Fixed Point Theorem is one of its basic results. Informally, it states that the least fixed point of a continuous operator can be computed by iterating the operator. It is heavily used in logic programming.

For the sake of self-containedness, this Appendix introduces the basic notions of Order Theory necessary to formally state and prove the Kleene Fixed Point Theorem. For an elaborate study of this topic with many further references, we refer the reader to \cite{Davey1990}.

The first definition is of a partially ordered set, under which we mean any set with an associated relation ``$\leq$'' that can be used to compare elements of this set. This relation is required to obey certain properties that can be naturally expected from any such ordering relation.

\begin{definition}[Partial Order]
A \emph{partial order} is a pair $(P, \leq)$ where $P$ is a set and $\leq$ is a reflexive, antisymetric and transitive relation over $P$, i.e. the following conditions are satisfied for all $a, b, c \in P$:
\begin{gather*}
a \leq a \\
(a \leq b \land b \leq a) \mlthen a = b \\
(a \leq b \land b \leq c) \mlthen a \leq c
\end{gather*}
We also say that $P$ is a partially ordered set (w.r.t. $\leq$).
\end{definition}

In logic programming, the set of interpretations usually forms a partial order that is usually ordered by the subset relation. In case of MKNF interpretations, the partial order is determined by the superset relation.

The following definitions introduce the least and greatest elements and lower and upper bounds of a subset of a partially ordered set.

\begin{definition}[Least and Greatest Element]
Let $P$ be a partially ordered set, $S \subseteq P$ and $a \in S$. Then $a$ is the \emph{least element of $S$} if for every $b \in S$ it holds that $a \leq b$, and $a$ is the \emph{greatest element of $S$} if for every $b \in S$ it holds that $b \leq a$.
\end{definition}

\begin{definition}[Lower and Upper Bound]
Let $P$ be a partially ordered set, $S \subseteq P$ and $a \in P$. Then $a$ is a \emph{lower bound of $S$} if for every $b \in S$ it holds that $a \leq b$, and $a$ is an \emph{upper bound of $S$} if for every $b \in S$ it holds that $b \leq a$.
\end{definition}

Combining the previous notions, we obtain the notion of a least upper bound (supremum) and greatest lower bound (infimum).

\begin{definition}[Supremum and Infimum]
Let $P$ be a partially ordered set, $S \subseteq P$ and $a \in P$. Then $a$ is the \emph{supremum of $S$}, denoted by $a = \sup(S)$, if it is the least element of the set of upper bounds of $S$, and $a$ is the \emph{infimum of $S$}, denoted by $a = \inf(S)$, if it is the greatest element of the set of lower bounds of $S$
\end{definition}

The next notion of a \emph{directed set} plays an important role in defining when a function on a partial order is continuous. It is also required in order to define a stricter structure than a partial order, the \emph{complete partial order}. We need to introduce both these notions in order to formulate the Kleene Fixed Point Theorem which describes one property of continuous functions on complete partial orders.

\begin{definition}[Directed Set]
A \emph{directed set} is a pair $(D, \leq)$ where $D$ is a non-empty set, $\leq$ is a reflexive and transitive relation over $D$ and for any elements $a, b \in D$ there exists some $c \in D$ such that $a \leq c$ and $b \leq c$.
\end{definition}

As can be seen, in a directed set, every pair of elements has an upper bound that also belongs to the set. This property can be naturally extended to finite subsets of the directed set.

\begin{proposition} \label{prop:order:dirset_upper_bound}
Let $(D, \leq)$ be a directed set and $S$ a finite subset of $D$. Then $D$ contains an upper bound of $S$.
\end{proposition}
\begin{proof}
Suppose $S = \set{s_1, s_2, \dotsc, s_n}$. Then we can construct a sequence $\{d_i\}_{i=2}^n$ of elements of $D$ such that
\begin{align*}
	& s_1 \leq d_2 \text{ and } s_2 \leq d_2 \enspace; \\
	& s_i \leq d_i \text{ and } d_{i-1} \leq d_i \qquad \text{for each } i \in \set{3, 4, \dotsc, n}.
\end{align*}
By induction on $i$ it follows that $d_i \leq d_n$ for every $i \in \set{2, 3, \ldots, n}$ and by applying transitivity we obtain $s_i \leq d_n$ for each $i \in \set{1, 2, \ldots, n}$. Hence $d_n$ is an upper bound of $S$ in $D$.
\end{proof}

As an important consequence, we obtain that every finite directed set contains its own supremum.

\begin{corollary} \label{cor:order:dirset_sup}
Any finite directed set contains its supremum.
\end{corollary}
\begin{proof}
Let $(D, \leq)$ be a finite directed set. Then by Prop. \ref{prop:order:dirset_upper_bound} it contains its own upper bound $d$. Consider some other upper bound $u$ of $D$. Then since $d \in D$, we have $d \leq u$ and so $d$ is the least upper bound of $D$, i.e. the supremum of $D$.
\end{proof}

We can now introduce two properties of functions on partial orders. the weaker property of monotonicity basically states that the function preserves the partial order:

\begin{definition}[Monotonic Function]
Let $P, Q$ be two partially ordered sets and $f:P \to Q$. We say $f$ is \emph{monotonic} if for every $a, b \in P$ such that $a \leq b$ we have $f(a) \leq f(b)$.
\end{definition}

The property of continuity is stricter and requires that for all directed sets with a supremum in the domain, the image of that supremum is the same as the supremum of images of elements of the directed set.

\begin{definition}[Continuous Function]
Let $P, Q$ be two partially ordered sets and $f:P \to Q$. We say $f$ is \emph{continuous} if for every directed subset $D$ of $P$ with supremum in $P$ it holds that
\[
\sup(f(D)) = f(\sup(D))
\]
where $f(A) = \set{f(a) | a \in A}$ for any set $A \subseteq P$.
\end{definition}

The next proposition formally proves that continuity is a stronger property thatn monotonicity.

\begin{proposition} \label{prop:order:cont_mono}
Every continuous function is monotonic.
\end{proposition}
\begin{proof}
Consider a continuous function $f:P \to Q$ and some $a, b \in P$ such that $a \leq b$. Then the set $D = \set{a, b}$ is a directed subset of $P$ and by continuity of $f$ we obtain
\[
\sup(f(D)) = f(\sup(D))
\]
Since $\sup(D) = b$, we further obtain $\sup(\set{f(a), f(b)}) = f(b)$ and consequently $f(a) \leq f(b)$ as desired.
\end{proof}

A complete partial is simply a partial order with a least element in which every directed set has a supremum. Many partially ordered structures, such as the space of interpretations, satisfy this property.

\begin{definition}[Complete Partial Order]
A partial order $(P, \leq)$ is a \emph{complete partial order} if $P$ has a least element and every directed subset $S$ of $P$ has a supremum in $P$.
\end{definition}

Finally, we are able to formulate and prove the main result of this appendix. It states that the least fixed point of a continuous function on a complete partial order always exists and can be approximated by iterations of the function applied to the least element of the complete partial order.

\begin{theorem}[Kleene Fixed Point Theorem] \label{thm:order:kleene}
Let $P$ be a complete partial order with the least element $\bot$ and $f$ be a continuous function on $P$. Then the least fixed point of $f$ is $\sup \set{f^n(\bot) | n \geq 0}$.
\end{theorem}
\begin{proof}
This is a well-established result, even so much that it is not easy to find its original source. The oldest source we were able to find and verify is the book \cite{Stoy1977}, pp. 112, Theorem 6.64. The same proof is also presented in the paper \cite{Stoy1979}, pp. 55 (according to the numbering of the Proceedings). A more recent book on this topic is \cite{Davey1990} where this result is formulated as Theorem 4.5 on pp. 89.

Now we start with the presentation of the proof. Suppose $f$ is a continuous function on the complete partial order $P$. Then by Proposition \ref{prop:order:cont_mono} it is monotonic from which it follows easily that the set $D = \set{f^n(\bot) | n \geq 0}$ is directed. Hence, its supremum $\sup D$ exists in $P$. We will now show that $\sup D$ is a fixed point of $f$:
\begin{align*}
f(\sup D) &= \sup f(D) = \sup f(\set{f^n(\bot) | n \geq 0}) = \sup \set{f^n(\bot) | n \geq 1} = \\
&= \sup ( \set{\bot} \cup \set{f^n(\bot) | n \geq 1} ) = \sup \set{f^n(\bot) | n \geq 0} = \\
&= \sup D
\end{align*}
Further, suppose $a$ is some fixed point of $f$. In order to prove that $\sup D$ is the least fixed point of $f$, we need to show that $\sup D \leq a$. By induction on $n$ we can easily obtain that $f^n(\bot) \leq a$ for all $n \geq 0$:
\renewcommand{\labelenumi}{\arabic{enumi}$^\circ$}
\begin{enumerate}
	\item $f^0(\bot) = \bot \leq a$
	\item By inductive assumption $f^{n-1}(\bot) \leq a$, so by monotonicity of $f$ we obtain $f^n(\bot) \leq f(a) = a$.
\end{enumerate}
So $a$ is an upper bound of $D$ and, by definition of a supremum, $\sup D \leq a$ as desired.
\end{proof}

\section{Properties of MKNF}

\subsection{General Properties}

\begin{lemma}[Models of Positive Sentences] \label{lemma:mknf:pos_s5}
Let $\phi$ be a positive MKNF sentence, $I$ a propositional interpretation and $M, N_0 \in \mint$. If $\mstr[I, M, N_0] \ent \phi$, then $\mstr \ent \phi$ for any $N \in \mint$.
\end{lemma}
\begin{proof}
Follows directly from Definition \ref{def:mknf:models} and the fact that the valuation of a positive formula in a structure $\mstr$ is independent of $N$.
\end{proof}

\begin{corollary} \label{cor:mknf:pos_mknf_s5}
Let $\phi$ be a positive MKNF sentence. Then the MKNF models of $\phi$ are exactly the subset-maximal S5 models of $\phi$.
\end{corollary}
\begin{proof}
Follows from Definition \ref{def:mknf:models} and Lemma \ref{lemma:mknf:pos_s5}.
\end{proof}

\begin{lemma}
Let $\leq$ be a binary relation defined on the set $\mint$ of all sets of first-order interpretations for any $M, N \in \mint$ as follows:
\[
M \leq N \mlequiv M \supseteq N
\]
Then $(\mint, \leq)$ is a complete partial order with the least element $\foint$.
\end{lemma}
\begin{proof}
Follows from the set-theoretic properties of the subset relation $\subseteq$ and of the set intersection $\cap$. Notice that even subsets of $\mint$ that are not directed have their supremum (intersection) in $\mint$.
\end{proof}

\begin{lemma} \label{lemma:mknf:mk_monotony}
Let $\phi$ be an first-order sentence and $M, N \in \mint$ be such that $M \leq N$. If $M \ent \mk \phi$, then also $N \ent \mk \phi$.
\end{lemma}
\begin{proof}
Suppose $M \ent \mk \phi$ and consider some interpretation $I \in N$. By the assumption we obtain $I \in M$ and so $\mstr[I, M, M] \ent \mk \phi$. Hence $\mstr[I, M, M] \ent \phi$ and since $\phi$ is a first-order formula, its valuation in the structure $\mstr[I, M, M]$ doesn't depend on $M$, so $\mstr[I, N, N] \ent \phi$. Furthermore, our choice of $I$ was arbitrary, so we can conclude that $\mstr[I, N, N] \ent \phi$ for all $I \in N$. Consequently, $N \ent \mk \phi$ as desired.
\end{proof}

\subsection{Models of First-Order Theories}

\begin{lemma}[Greatest Model of a First-Order Theory] \label{lemma:mknf:fo model}
For any first-order theory $\thr$ it holds that
\[
\smod(\thr) = \Set{ I \in \foint | (\forall \phi \in \thr)(I \ent \phi) }
\]
\end{lemma}
\begin{proof}
We will prove that
\[
M_{\thr} = \Set{ I \in \foint | (\forall \phi \in \thr)(I \ent \phi) }
\]
is the greatest set among the sets $M \in \mint$ with the property $M \ent \thr$.

First we need to prove that $M_{\thr}$ satisfies this property, i.e. that $M_{\thr} \ent \thr$. Take some $\phi \in \thr$ and $I \in M_{\thr}$. Then $I \ent \phi$ and since $\phi$ is first-order, we also obtain $\mstr[I, M_{\thr}, M_{\thr}] \ent \phi$. This holds for any $I \in M_{\thr}$, so $M_{\thr} \ent \phi$.

Now let $M \in \mint$ be such that $M \ent \thr$ and suppose $I \in M$. Then for every $\phi \in \thr$ we must have $\mstr[I, M, M] \ent \phi$ and since $\phi$ is first-order, this entails $I \ent \phi$. Hence, $I \in M_{\thr}$, so $M \subseteq M_S$. This fact finishes our proof.
\end{proof}

\subsection{Relevant Part of an MKNF Interpretation}

\begin{definition}[Predicate Symbols Relevant to a Ground Formula]
Given a ground MKNF formula $\phi$, we define the \emph{set $\lpref$ of predicate symbols relevant to $\phi$} inductively as follows:
\renewcommand{\labelenumi}{\arabic{enumi}$^\circ$}
\begin{enumerate}
	\item If $\phi$ is a first-order atom $P(c_1, c_2, \dotsc, c_n)$, then $\lpref = \set{P}$;
	\item If $\phi$ is of the form $\lnot \psi$, then $\lpref = \lpref[\psi]$;
	\item If $\phi$ is of the form $\phi_1 \land \phi_2$, then $\lpref = \lpref[\phi_1] \cup \lpref[\phi_2]$;
	\item If $\phi$ is of the form $\mk \psi$, then $\lpref = \lpref[\psi]$;
	\item If $\phi$ is of the form $\mnot \psi$, then $\lpref = \lpref[\psi]$.
\end{enumerate}
\end{definition}

\begin{definition}[Constant Symbols Relevant to a Ground Formula]
Given a ground MKNF formula $\phi$, we define the \emph{set $\lconf$ of constant symbols relevant to $\phi$} inductively as follows:
\renewcommand{\labelenumi}{\arabic{enumi}$^\circ$}
\begin{enumerate}
	\item If $\phi$ is a first-order atom $P(c_1, c_2, \dotsc, c_n)$, then $\lconf = \set{c_1, c_2, \dotsc, c_n}$;
	\item If $\phi$ is of the form $\lnot \psi$, then $\lconf = \lconf[\psi]$;
	\item If $\phi$ is of the form $\phi_1 \land \phi_2$, then $\lconf = \lconf[\phi_1] \cup \lconf[\phi_2]$;
	\item If $\phi$ is of the form $\mk \psi$, then $\lconf = \lconf[\psi]$;
	\item If $\phi$ is of the form $\mnot \psi$, then $\lconf = \lconf[\psi]$.
\end{enumerate}
\end{definition}

\begin{definition}[Restriction of an MKNF Interpretation]
Let $I \in \foint$ and $M \in \mint$. Given a finite set of predicate symbols $\lpre' \subseteq \lpre$ and a set of constant symbols $\lcon' \subseteq \Delta$, we define the \emph{restriction of $I$ to $\lpre'$ and $\lcon'$} as the Herbrand first-order interpretation $\restr{I}$ over the Herbrand Universe $\lcon'$ that interpretes only the predicates from $\lpre'$ in such a way that
\[
(c_1, c_2, \dotsc, c_n) \in P^{\restr{I}} \mlequiv (c_1, c_2, \dotsc, c_n) \in P^I
\]
where $P \in \lpre'$ and $c_1, c_2, \dotsc, c_n \in \lcon'$. We also define the \emph{restriction of $M$ to $\lpre'$ and $\lcon'$} as $\restr{M} = \Set{\restr{I} | I \in M}$.
\end{definition}

\begin{lemma}[Truth of Ground Formulas under Restriction to Relevant Symbols] \label{lemma:mknf:relevant}
Let $\phi$ be a ground MKNF formula, $\lpre' \subseteq \lpre$ a finite set of predicate symbols such that $\lpre' \supseteq \lpref$, $\lcon' \subseteq \Delta$ a finite set of constant symbols such that $\lcon' \supseteq \lconf$, $I$ a propositional interpretation and $M, N \in \mint$. Then
\[
\mstr \ent \phi \mlequiv \mstr[\restr{I}, \restr{M}, \restr{N}] \ent \phi \enspace.
\]
\end{lemma}
\begin{proof*} \label{proof:mknf:ent_relevant}
We will prove by structural induction on $\phi$:
\renewcommand{\labelenumi}{\arabic{enumi}$^\circ$}
\begin{enumerate}
	\item If $\phi$ is a ground first-order atom of the form $P(c_1, c_2, \dotsc, c_n)$, then $P \in \lpref$ and $c_1, c_2, \dotsc, c_n \in \lconf$, so $P \in \lpre'$ and $c_1, c_2, \dotsc, c_n \in \lcon'$. The following chain of equivalences now proves the claim:
	\begin{align*}
	\mstr \ent \phi &\mlequiv (c_1, c_2, \dotsc, c_n) \in P^I \mlequiv (c_1, c_2, \dotsc, c_n) \in P^{\restr{I}} \\
	& \mlequiv \mstr[\restr{I}, \restr{M}, \restr{N}] \ent \phi \enspace;
	\end{align*}
	
	\item If $\phi$ is of the form $\lnot \psi$, then $\lpref = \lpref[\psi]$ and $\lconf = \lconf[\psi]$, so $\lpre' \supseteq \lpref[\psi]$ and $\lcon' \supseteq \lconf[\psi]$. Hence, we can use the inductive hypothesis for $\psi$ as follows:
	\begin{align*}
	\mstr \ent \phi &\mlequiv \mstr \nent \psi \mlequiv \mstr[\restr{I}, \restr{M}, \restr{N}] \nent \psi \\
	& \mlequiv \mstr[\restr{I}, \restr{M}, \restr{N}] \ent \phi \enspace;
	\end{align*}
	
	\item If $\phi$ is of the form $\phi_1 \land \phi_2$, then $\lpref = \lpref[\phi_1] \cup \lpref[\phi_2]$ and $\lconf = \lconf[\phi_1] \cup \lconf[\phi_2]$ and we can easily verify that the inductive assumption can be used on both $\phi_1$ and $\phi_2$ and the proposition can be proved for $\phi$ as follows:
	\begin{align*}
	\mstr \ent \phi &\mlequiv \mstr \ent \phi_1 \land \mstr \ent \phi_2 \\
	&\mlequiv \mstr[\restr{I}, \restr{M}, \restr{N}] \ent \phi_1 \\
		& \qquad \qquad \qquad {}\land \mstr[\restr{I}, \restr{M}, \restr{N}] \ent \phi_2 \\
	&\mlequiv \mstr[\restr{I}, \restr{M}, \restr{N}] \ent \phi \enspace;
	\end{align*}
	
	\item If $\phi$ is of the form $\mk \psi$, then $\lpref = \lpref[\psi]$ and $\lconf = \lconf[\psi]$, so $\lpre' \supseteq \lpref[\psi]$ and $\lcon' \supseteq \lconf[\psi]$. The claim now follows from the inductive hypothesis for $\psi$:
	\begin{align*}
	\mstr \ent \phi &\mlequiv \br{\forall J \in M}\br{\mstr[J, M, N] \ent \psi} \\
	&\mlequiv \br{\forall J \in M}\br{\mstr[\restr{J}, \restr{M}, \restr{N}] \ent \psi} \\
	&\mlequiv \br{\forall J \in \restr{M}}\br{\mstr[J, \restr{M}, \restr{N}] \ent \psi} \\
	&\mlequiv \mstr[\restr{I}, \restr{M}, \restr{N}] \ent \phi \enspace;
	\end{align*}
	
	\item If $\phi$ is of the form $\mnot \psi$, then $\lpref = \lpref[\psi]$ and $\lconf = \lconf[\psi]$, so $\lpre' \supseteq \lpref[\psi]$ and $\lcon' \supseteq \lconf[\psi]$. The claim follows similarly as in the previous case:
	\begin{align*}
	\mstr \ent \phi &\mlequiv \br{\exists J \in N}\br{\mstr[J, M, N] \nent \psi} \\
	&\mlequiv \br{\exists J \in N}\br{\mstr[\restr{J}, \restr{M}, \restr{N}] \nent \psi} \\
	&\mlequiv \br{\exists J \in \restr{N}}\br{\mstr[J, \restr{M}, \restr{N}] \nent \psi} \\
	&\mlequiv \mstr[\restr{I}, \restr{M}, \restr{N}] \ent \phi \enspace. \qedhere
	\end{align*}
\end{enumerate}
\end{proof*}

\section{Properties of Hybrid Knowledge Bases}

\begin{lemma} \label{lemma:mknf programs:modal atoms}
Let $\mathcal{Z}$ be a set of first-order theories. Then
\[
\smod \left( \bigcup \mathcal{Z} \right) = \bigcap \smod(\mathcal{Z})
\]
where $\smod(\mathcal{Z}) = \Set{ \smod(\thr) | \thr \in \mathcal{Z} }$.
\end{lemma}
\begin{proof*}
The following sequence of equivalences proves the claim:
\begin{align*}
I \in \smod \left( \bigcup \mathcal{Z} \right)
&\mlequivbylemma{\text{Lemma \ref{lemma:mknf:fo model}}}
	\left( \forall \phi \in \bigcup \mathcal{Z} \right) (I \ent \phi) \\
&\mlequivbylemma{}
	(\forall \thr \in \mathcal{Z}) (\forall \phi \in \thr) (I \ent \phi) \\
&\mlequivbylemma{\text{Lemma \ref{lemma:mknf:fo model}}}
	(\forall \thr \in \mathcal{Z}) (I \in \smod(\thr)) \\
&\mlequivbylemma{}
	I \in \bigcap \smod(\mathcal{Z}) \qedhere
\end{align*}
\end{proof*}

\begin{definition}[Hybrid Immediate Consequence Operator]
The \emph{immediate consequence operator} associated with the definite $\prog$-ground hybrid knowledge base $\kb = \an{\ont, \prog}$ is a mapping $\tkb: \mint \to \mint$ defined for any $M \in \mint$ as
\[
\tkb(M) = \smod( \ont \cup \Set{ H^*(r) | r \in \prog \land M \ent B(r)} )
\]
\end{definition}

\begin{lemma} \label{lemma:mknf programs:tkb alt}
Let $\kb = \an{\ont, \prog}$ be definite $\prog$-ground hybrid knowledge base. Then for every $M \in \mint$ it holds that
\[
\tkb(M) = \smod(\ont) \cap \smod( \Set{ H^*(r) | r \in \prog \land M \ent B(r)} )
\]
\end{lemma}
\begin{proof}
Let $\thr = \Set{ H^*(r) | r \in \prog \land M \ent B(r)}$. We need to show that
\[
\smod(\ont \cup \thr) = \smod(\ont) \cap \smod(\thr) \enspace.
\]
This follows from Lemma \ref{lemma:mknf programs:modal atoms}.
\end{proof}

\begin{lemma} \label{lemma:dirset_body}
Let $\mathcal{D}_F$ be a finite directed set of first-order interpretations and $r$ be a ground definite rule. Then
\[
\bigcap \mathcal{D}_F \ent B(r) \mlequiv (\exists M \in \mathcal{D}_F)(M \ent B(r))
\]
\end{lemma}
\begin{proof}
By Corollary \ref{cor:order:dirset_sup} we have $\bigcap \mathcal{D}_F \in \mathcal{D}_F$, so if $\bigcap \mathcal{D}_F \ent B(r)$, then also $(\exists M \in \mathcal{D}_F)(M \ent B(r))$. Now suppose that $M \ent B(r)$ for some $M \in \mathcal{D}_F$. Then $M \leq \bigcap \mathcal{D}_F$ and by a repeated use of Lemma \ref{lemma:mknf:mk_monotony} for each conjunct of $B(r)$ we obtain $\bigcap \mathcal{D}_F \ent B(r)$.
\end{proof}

\begin{lemma} \label{lemma:restrcapd_caprestrd}
Let $\mathcal{D}$ be a directed set of MKNF interpretations, $\lpre'$ a set of predicate symbols and $\lcon'$ a set of constant symbols and
\begin{equation} \label{eq:proof:da}
\restr{\mathcal{D}} = \Set{\restr{M} | M \in \mathcal{D}}
\end{equation}
Then the following holds:
\[
\restr{\left( \bigcap \mathcal{D} \right)} = \bigcap \restr{\mathcal{D}}
\]
\end{lemma}
\begin{proof*}
\begin{align*}
\restr{\left( \bigcap \mathcal{D} \right)}
&= \restr{\left( \bigcap \Set{M | M \in \mathcal{D}} \right)} \\
&= \restr{\left( \Set{I | (\forall M \in \mathcal{D})(I \in M)} \right)} \\
&= \left( \Set{ \restr{I} | (\forall M \in \mathcal{D})(I \in M)} \right) \\
&= \left( \Set{I | (\forall M \in \mathcal{D}) \left( I \in \restr{M} \right)} \right) \\
&= \left( \Set{I | \left( \forall M \in \restr{\mathcal{D}} \right)(I \in M)} \right) \\
&= \bigcap \Set{M | M \in \restr{\mathcal{D}}}
	= \bigcap \restr{\mathcal{D}} \qedhere
\end{align*}
\end{proof*}

\begin{lemma} \label{lemma:capd}
Let $\mathcal{D}$ be a directed set of MKNF interpretations and $r$ a ground definite rule. Then:
\[
(\exists M \in \mathcal{D})(M \ent B(r)) \mlequiv \bigcap \mathcal{D} \ent B(r)
\]
\end{lemma}
\begin{proof*}
Let $\lpre' = \lpref[B(r)]$ and $\lcon' = \lconf[B(r)]$ and consider these equivalences:
\begin{align*}
(\exists M \in \mathcal{D})(M \ent B(r))
&\mlequivbylemma{\text{Lemma \ref{lemma:mknf:relevant}}}
	(\exists M \in \mathcal{D}) \left( \restr{M} \ent B(r) \right) \\
&\mlequivbylemma{\text{\eqref{eq:proof:da}}}
	\left( \exists M \in \restr{\mathcal{D}} \right) (M \ent B(r)) \\
&\mlequivbylemma{\text{Lemma \ref{lemma:dirset_body}}}
	\left( \bigcap \restr{\mathcal{D}} \right) \ent B(r) \\
&\mlequivbylemma{\text{Lemma \ref{lemma:restrcapd_caprestrd}}}
	\restr{\left( \bigcap \mathcal{D} \right)} \ent B(r) \\
&\mlequivbylemma{\text{Lemma \ref{lemma:mknf:relevant}}}
	\bigcap \mathcal{D} \ent B(r) \qedhere
\end{align*}
\end{proof*}

\begin{proposition}[Continuity of $\tkb$] \label{prop:mknf_programs:tp_cont}
Let $\kb = \an{\ont, \prog}$ be a definite $\prog$-ground hybrid knowledge base. Then $\tkb$ is a continuous function on $\mint$.
\end{proposition}
\begin{proof*}
Consider some directed subset $\mathcal{D}$ of $\mint$. To prove that $\tkb$ is continuous, we need to show that $\sup(\tkb(\mathcal{D})) = \tkb(\sup(\mathcal{D}))$.
By Lemma \ref{lemma:mknf programs:tkb alt}, we have:
\[
\sup(\tkb(\mathcal{D})) = \smod(\ont) \cap
	\bigcap_{M \in \mathcal{D}} \smod \left( \Set{ H^*(r) | r \in \prog \land M \ent B(r)} \right) \enspace.
\]
Let $S$ denote the set
\begin{equation} \label{eq:proof:S as shortcut:1}
\bigcap_{M \in \mathcal{D}} \smod \left( \Set{ H^*(r) | r \in \prog \land M \ent B(r)} \right)
\end{equation}
so that
\begin{equation} \label{eq:proof:S as shortcut:2}
\sup(\tkb(\mathcal{D})) = \smod(\ont) \cap S
\end{equation}
Consider the following identities:
\begin{align*}
S
&\eqbylemma[\text{Lemma \ref{lemma:mknf programs:modal atoms}}]
	 \smod \left( \bigcup_{M \in \mathcal{D}} \Set{ H^*(r) | r \in \prog \land M \ent B(r)} \right) \\
&\eqbylemma
	\smod \left( \Set{ H^*(r) | r \in \prog \land (\exists M \in \mathcal{D})(M \ent B(r))} \right) \\
&\eqbylemma[\text{Lemma \ref{lemma:capd}}]
	\smod \left( \Set{ H^*(r) | r \in \prog \land \bigcap \mathcal{D} \ent B(r)} \right)
\end{align*}
Together with \eqref{eq:proof:S as shortcut:2} and Lemma \ref{lemma:mknf programs:tkb alt} this implies that
\begin{align*}
\sup(\tkb(\mathcal{D})) &= \smod(\ont) \cap S \\
	&= \smod(\ont) \cap \smod \left( \Set{ H^*(r) | r \in \prog \land \bigcap \mathcal{D} \ent B(r)} \right) \\
	&= \tkb(\sup(\mathcal{D})) \enspace. \qedhere
\end{align*}
\end{proof*}

\begin{corollary}[Monotonicity of $\tkb$] \label{cor:mknf_programs:tp_mono}
Let $\kb = \an{\ont, \prog}$ be a definite $\prog$-ground hybrid knowledge base. Then $\tkb$ is a monotonic function on $\mint$ and for any $n \geq 0$ it holds that $\tkb^n(\foint) \supseteq \tkb^{n+1}(\foint)$.
\end{corollary}
\begin{proof}
The monotonicity of $\tkb$ follows directly from Props. \ref{prop:mknf_programs:tp_cont} and \ref{prop:order:cont_mono}. Now since $\foint$ is the minimal element of $(\mint, \leq)$, we obtain $\tkb^0(\foint) = \foint \leq \tkb^1(\foint)$. By $n$ times applying the monotonicity of $\tkb$ we obtain $\tkb^n(\foint) \leq \tkb^{n+1}(\foint)$ which is equivalent to $\tkb^n(\foint) \supseteq \tkb^{n+1}(\foint)$.
\end{proof}

The following proposition shows that each definite $\prog$-ground hybrid knowledge base either has no model at all, or, similarly as definite logic programs, it has the greatest S5 model that coincides with its unique MKNF model. It also shows how this model can be computed by iterating the $\tkb$ operator starting from $\foint$.

\begin{proposition} \label{prop:mknf_programs:least_model}
Let $\kb = \an{\ont, \prog}$ be a definite $\prog$-ground hybrid knowledge base. Then either $\kb$ has no S5 model or it has the greatest S5 model that also coincides with its single MKNF model. Furthermore, the set
\[
\smod(\kb) = \bigcap_{n \geq 0} \tkb^n(\foint)
\]
is empty if $\kb$ has no S5 model and otherwise coincides with its unique MKNF model.
\end{proposition}
\begin{proof} \label{proof:mknf_programs:least_model}
First we will prove an auxiliary claim: $M \subseteq \tkb(M)$ holds for any S5 model $M$ of $\kb$. Suppose $M$ is an S5 model of $\kb$ and recall that
\[
\tkb(M) = \smod \left( \ont \cup \Set{ H^*(r) | r \in \prog \land M \ent B(r)} \right)
\]
Let's take some formula $\phi \in \ont$. We know that $M \ent \phi$ because $M$ is an S5 model of $\kb$. Now consider some rule $r \in \prog$ such that $M \ent B(r)$. Since $M$ is an S5 model of $\kb$, we obtain $M \ent H^*(r)$. Consequently, $M \ent H^*(r)$ for every such $r$. So $M$ is an S5 model of $\ont \cup \Set{ H^*(r) | r \in \prog \land M \ent B(r)}$ and since $\tkb(M)$ is by definition of $\smod(\cdot)$ the greatest S5 model of $\ont \cup \Set{ H^*(r) | r \in \prog \land M \ent B(r)}$, we can conclude that $M \subseteq \tkb(M)$.

Now we will proceed with the main part of the proof. Let
\[
M_{\kb} = \bigcap_{n \geq 0} \tkb^n(\foint)
\]
Then, by Corollary \ref{prop:mknf_programs:tp_cont} and Theorem \ref{thm:order:kleene}, $M_{\kb}$ is the least fixed point of $\tkb$. First we will show that $M_{\kb}$ contains every S5 model of $\kb$. Assume, to the contrary, that $M$ is an S5 model of $\kb$ such that $M \nsubseteq M_{\kb}$. Then by definition $M \subseteq \foint = \tkb^0(\foint)$. It cannot be the case that $M \subseteq \tkb^n(\foint)$ for all $n \geq 0$ because that would be in conflict with $M \nsubseteq M_{\kb}$. So let
\[
n_0 = \max \set{n \geq 0 | M \subseteq \tkb^n(\foint)} \enspace.
\]
Now we have $M \subseteq \tkb^{n_0}(\foint)$ and by the auxiliary claim proved above, we obtain $M \subseteq \tkb(M)$ which together with the monotonicity of $\tkb$ (Corollary \ref{cor:mknf_programs:tp_mono}) yields $M \subseteq \tkb(M) \subseteq \tkb(\tkb^{n_0}(\foint)) = \tkb^{n_0+1}(\foint)$. However, this is in conflict with the definition of $n_0$, so no S5 model $M$ of $\kb$ with $M \nsubseteq M_{\kb}$ can exist.

Now we will show that $M_{\kb}$ models $\kb$. This can be easily verified for every $\phi \in \ont$. Take some $r \in \prog$. If $M_{\kb} \nent B(r)$, then $M_{\kb} \ent r$ and we are done. So assume $M_{\kb} \ent B(r)$. In this case we can use the fixpoint property of $M_{\kb}$:
\[
M_{\kb} = \tkb ( M_{\kb} ) = \smod ( \ont \cup \Set{H^*(r) | r \in \prog \land M_{\kb} \ent B(r)} )
\]
and conclude that $M_{\kb} \ent H^*(r)$. Consequently also $M_{\kb} \ent r$.

We already proved that $M_{\kb}$ is the greatest set of interpretations that models  $\kb$. So in case $\kb$ has no S5 model, $M_{\kb}$ will be empty. On the other hand, if $\kb$ has some S5 model, this model is included in $M_{\kb}$, so $M_{\kb}$ is non-empty and hence is the greatest S5 model of $\kb$. Further, by Corollary \ref{cor:mknf:pos_mknf_s5} it follows that $M_{\kb}$ is also the unique MKNF model of $\kb$.
\end{proof}

For MKNF models of arbitrary $\prog$-ground hybrid knowledge bases we also obtain a characterisation that is similar to the fixpoint definition of stable models of normal logic programs:

\begin{proposition} \label{prop:mknf_programs:mknf_fixpoint}
An MKNF interpretation $M$ is an MKNF model of a $\prog$-ground hybrid knowledge base $\kb = \an{\ont, \prog}$ if and only if $M = \smod(\an{\ont, \progmod})$ where
\[
\progmod = \Set{ H(r) \lif B^+(r) | r \in \prog \land M \ent B^-(r) }
\]
\end{proposition}
\begin{proof} \label{proof:mknf_programs:mknf_fixpoint}
First notice that since $\prog$ is ground, $\pi(r) = r$ for every $r \in \prog \cup \progmod$.

Let $\kb^M = \an{\ont, \progmod}$ and suppose $M$ is an MKNF model of $\kb$. First we will show that $M$ is an S5 model of $\kb^M$. Obviously, $M$ models all formulas from $\pi(\ont)$.  Suppose that $r^M = (H(r) \lif B^+(r))$ is a rule from $\progmod$. If $M \nent B^+(r)$, then $M \ent r^M$. On the other hand, if $M \ent B^+(r)$, then $M \ent r$ implies also $M \ent H(r)$. Consequently, $M \ent r^M$.

As $M$ is an S5 model of $\kb^M$, it must hold that $M$ is a subset of $\smod(\kb^M)$ because $\smod(\kb^M)$ is the greatest S5 model of $\kb^M$. By contradiction, we will show that $M = \smod(\kb^M)$. Assume $M \subsetneq \smod(\kb^M)$. Since $M$ is an MKNF model of $\kb$, there must be some formula $\phi \in \pi(\kb)$ and some $I' \in \smod(\kb^M)$ such that $\mstr[I', \smod(\kb^M), M] \nent \phi$. But $\smod(\kb^M)$ models $\pi(\ont)$, so $\phi$ must be some rule $r$ from $\prog$ and the following must hold
\begin{multline*}
\mstr[I', \smod(\kb^M), M] \ent B^-(r) \land \mstr[I', \smod(\kb^M), M] \ent B^+(r) \\ \land \mstr[I', \smod(\kb^M), M] \nent H(r)
\end{multline*}
which is equivalent to
\[
M \ent B^-(r) \land \smod(\kb^M) \ent B^+(r) \land \smod(\kb^M) \nent H(r) \enspace.
\]
However, this is in conflict with $\smod(\kb^M)$ being an S5 model of $\kb^M$ since $H(r) \lif B^+(r) \in \progmod$.

For the converse implication, assume $M$ is an MKNF interpretation such that $M = \smod(\kb^M)$. It must hold that $M \ent \pi(\ont)$, so consider some rule $r \in \prog$. If $M \nent B^-(r)$, then $M$ is trivially a model of $r$. On the other hand, if $M \ent B^-(r)$, then $M$ is also a model of $H(r) \lif B^+(r)$, so $M$ is again a model of $r$. Consequently, $M$ is an S5 model of $\kb$. Now take some $M' \supsetneq M$. Then since $M$ is the greatest model of $\kb^M$, there is some rule $r^M = (H(r) \lif B^+(r)) \in \progmod$ such that $M' \nent r^M$, i.e.
\[
M \ent B^-(r) \land M' \ent B^+(r) \land M' \nent H(r)
\]
For any $I' \in M'$, this is equivalent to
\[
\mstr[I', M', M] \ent B^-(r) \land \mstr[I', M', M] \ent B^+(r) \land \mstr[I', M', M] \nent H(r)
\]
which in turn is equivalent to $\mstr[I', M', M] \nent r$. So $M$ is indeed an MKNF model of $\kb$.
\end{proof}

\section{Properties of the Hybrid Update Operator} \label{app:hybrid update operator}

\begin{proposition*}{prop:hybrid update operator:tpu continuous}
Let $\prog$ be a finite ground definite program, $\tbox$ a TBox and $\abox$ a sequence of ABoxes. Then $T_{\prog \oplus^{\tbox} \abox}$ is a continuous function on the complete partial order of all subsets of $\foint$ with the least element $\foint$.
\end{proposition*}
\begin{proof}[Proof of Proposition \ref{prop:hybrid update operator:tpu continuous}]
\label{proof:hybrid update operator:tpu continuous}
Consider some directed subset $\mathcal{D}$ of $\mint$. To prove that $T_{\prog \oplus^{\tbox} \abox}$ is continuous, we need to show that
\[
\sup(T_{\prog \oplus^{\tbox} \abox}(\mathcal{D})) = T_{\prog \oplus^{\tbox} \abox}(\sup(\mathcal{D})) \enspace.
\]
To simplify notation in this proof, we define for any set of first-order interpretations $M$ the following set:
\[
\mathsf{con}(M) = \Set{ H^*(r) | r \in \prog \land M \ent B(r) }
\]
Notice that if $M \supseteq N$ (or $M \leq N$ using the partial order on sets of first-order intepretations), then $\mathsf{con}(M) \subseteq \mathsf{con}(N)$.

By definition we now have
\begin{align}
\begin{split}
	T_{\prog \oplus^{\tbox} \abox}(\sup(\mathcal{D}))
	&= \smod \left( \mathsf{con} \left( \bigcap \mathcal{D} \right) \oplus^{\tbox} \abox \right) \\
	&= \mathsf{incorporate}^{\tbox} \left( \abox, \smod \left( \tbox \cup \mathsf{con} \left( \bigcap \mathcal{D} \right) \right) \right) \\
	&= \mathsf{incorporate}^{\tbox} \left( \abox, \smod(\tbox) \cap \smod \left( \mathsf{con} \left( \bigcap \mathcal{D} \right) \right) \right)
\end{split} \label{eq:proof:hybrid update operator:tpu continuous:1} \\
\intertext{and}
\begin{split}
	\sup(T_{\prog \oplus^{\tbox} \abox}(\mathcal{D}))
	&= \bigcap_{M \in \mathcal{D}} \smod \left( \mathsf{con}(M) \oplus^{\tbox} \abox \right) \\
	&= \bigcap_{M \in \mathcal{D}} \mathsf{incorporate}^{\tbox} \left( \abox, \smod(\tbox \cup \mathsf{con}(M)) \right) \\
	&= \bigcap_{M \in \mathcal{D}} \mathsf{incorporate}^{\tbox} \left( \abox, \smod(\tbox) \cap \smod(\mathsf{con}(M)) \right)
\end{split} \label{eq:proof:hybrid update operator:tpu continuous:2}
\end{align}
First suppose that a first-order interpretation $I$ is in $T_{\prog \oplus^{\tbox} \abox}(\sup(\mathcal{D}))$. Then by the previous equation we have that there is some $J \in \smod(\tbox) \cap \smod(\mathsf{con}(\bigcap \mathcal{D}))$ such that
\[
I \in \mathsf{incorporate}^{\tbox}(\abox, J) \enspace.
\]
Further, for every $M \in \mathcal{D}$ it holds that $\mathsf{con}(M) \subseteq \mathsf{con}(\bigcap \mathcal{D})$, and, hence, also that $\smod(\mathsf{con}(\bigcap \mathcal{D})) \subseteq \smod(\mathsf{con}(M))$. Consequently, $J \in \smod(\mathsf{con}(M))$ for every $M \in \mathcal{D}$, and so
\[
I \in \mathsf{incorporate}^{\tbox} \left( \abox, \smod(\tbox) \cap \smod(\mathsf{con}(M)) \right)
\]
also holds for every $M \in \mathcal{D}$. By \eqref{eq:proof:hybrid update operator:tpu continuous:2} we can now conclude that
\[
I \in \sup(T_{\prog \oplus^{\tbox} \abox}(\mathcal{D})) \enspace.
\]

For the converse inclusion, suppose $I \notin T_{\prog \oplus^{\tbox} \abox}(\sup(\mathcal{D}))$ and let $S$ be the set of all first-order interpretations $J \in \smod(\tbox)$ such that
\[
I \in \mathsf{incorporate}^{\tbox}(\abox, J) \enspace.
\]
By \eqref{eq:proof:hybrid update operator:tpu continuous:1} we obtain that $S \cap \smod \left( \mathsf{con} \left( \bigcap \mathcal{D} \right) \right) = \emptyset$, i.e. that each $J \in S$ is not a model of some atom $p_J$ such that there is a rule $r_J \in \prog$ with $H(r_J) = \mk p_J$ and $\bigcap \mathcal{D} \ent B(r_J)$. By Lemma \ref{lemma:capd}, this implies that for some $M \in \mathcal{D}$ we also have $M \ent B(r_J)$. Further, there are only finitely many rules in $\prog$, so by the directedness of $\mathcal{D}$ we can find an interpretation $M_S \in \mathcal{D}$ such that $M_S \ent B(r_J)$ for all $J \in S$. For this interpretation it will hold that $S \cap \smod(\mathsf{con}(M_S)) = \emptyset$. Hence,
\[
I \notin \mathsf{incorporate}^{\tbox}(\abox, \smod(\tbox) \cap \smod(\mathsf{con}(M_S)))
\]
and by \eqref{eq:proof:hybrid update operator:tpu continuous:2} we obtain that $I \notin \sup(T_{\prog \oplus^{\tbox} \abox}(\mathcal{D}))$.
\end{proof}

\begin{proposition*}{prop:hybrid update operator:KM:1}
Let $\prog$ be a finite ground program, $\tbox$ a TBox, $\abox$ an ABox and $M$ a minimal change dynamic stable model of $\prog \oplus^{\tbox} \abox$. Then $M \ent \abox$.
\end{proposition*}
\begin{proof}[Proof of Proposition \ref{prop:hybrid update operator:KM:1}]
\label{proof:hybrid update operator:KM:1}
If $M$ is a minimal change dynamic stable model of $\prog \oplus^{\tbox} \abox$, then it is a fixed point of $T_{\progmod \oplus^{\tbox} \abox}$, i.e.
\[
M = T_{\progmod \oplus^{\tbox} \abox}(M) = \smod \left( \Set{ H^*(r) | r \in \progmod \land M \ent B(r) } \oplus^{\tbox} \abox \right)
\]
and by the definition of the classical minimal change update operator it must hold that every $I \in M$ is a model of $\abox$. In other words, $M \ent \abox$.
\end{proof}

\begin{proposition*}{prop:hybrid update operator:KM:4}
Let $\prog$ be a finite ground program, $\tbox, \tbox'$ be TBoxes such that $\smod(\tbox) = \smod(\tbox')$, $\abox, \abox'$ be ABoxes such that $\smod(\abox) = \smod(\abox')$ and $M$ be an MKNF interpretation. Then $M$ is a minimal change dynamic stable model of $\prog \oplus^{\tbox} \abox$ if and only if $M$ is a minimal change dynamic stable model of $\prog \oplus^{\tbox'} \abox'$.
\end{proposition*}
\begin{proof}[Proof of Proposition \ref{prop:hybrid update operator:KM:4}]
\label{proof:hybrid update operator:KM:4}
Follows from the fact that the operators $T_{\progmod \oplus^{\tbox} \abox}$ and $T_{\progmod \oplus^{\tbox'} \abox'}$ are identical because the classical minimal change update operator only operates with models of $\tbox$, $\tbox'$, $\abox$ and $\abox'$, and not with their syntactic representation.
\end{proof}

\begin{proposition*}{prop:hybrid update operator:mknf}
Let $\prog$ be a finite ground program, $\ont = \tbox \cup \abox$ an ontology with TBox $\tbox$ and ABox $\abox$ and $M$ an MKNF interpretation such that for every subset $S$ of the set $\set{ H^*(r) | r \in \prog \land M \ent B(r) }$ the following condition is satisfied:
\[
\smod( S \oplus^{\tbox} \abox ) = \smod( S \cup \ont ) \enspace.
\]
Then $M$ is an MKNF model of $\an{\ont, \prog}$ if and only if $M$ is a minimal change dynamic stable model of $\prog \oplus^{\tbox} \abox$.
\end{proposition*}
\begin{proof}[Proof of Proposition \ref{prop:hybrid update operator:mknf}]
\label{proof:hybrid update operator:mknf}
By Propositions \ref{prop:mknf_programs:mknf_fixpoint} and \ref{prop:mknf_programs:least_model}, $M$ is an MKNF model of $\an{\ont, \prog}$ if and only if
\begin{equation} \label{eq:proof:hybrid update operator:mknf:mknf model}
M = \bigcap_{n \geq 0} T_{\an{\ont, \progmod}}^n(\foint)
\end{equation}
where for any set of first-order interpretations $N$ we have
\[
T_{\an{\ont, \progmod}}(N) = \smod \left( \ont \cup \Set{ H^*(r) | r \in \progmod \land N \ent B(r) } \right) \enspace.
\]
On the other hand, by Proposition \ref{prop:hybrid update operator:tpu continuous} and Theorem \ref{thm:order:kleene}, $M$ is a minimal change dynamic stable model of $\prog \oplus^{\tbox} \abox$ if and only if
\begin{equation} \label{eq:proof:hybrid update operator:mknf:dynamic model}
M = \bigcap_{n \geq 0} T_{\progmod \oplus^{\tbox} \abox}^n(\foint)
\end{equation}
where for any set of first-order interpretations $N$ we have
\[
T_{\progmod \oplus^{\tbox} \abox}(N) = \smod \left( \Set{ H^*(r) | r \in \progmod \land N \ent B(r) } \oplus^{\tbox} \abox \right) \enspace.
\]

Suppose now that $M$ is an MKNF model of $\an{\ont, \prog}$. Then from \eqref{eq:proof:hybrid update operator:mknf:mknf model} and Lemma \ref{lemma:mknf:mk_monotony} we obtain that for every $n \in \nat$ that
\[
\Set{ H^*(r) | r \in \progmod \land T_{\an{\ont, \progmod}}^n(\foint) \ent B(r) } \subseteq \Set{ H^*(r) | r \in \prog \land M \ent B(r) } \enspace.
\]
Hence, by the assumption of the proposition,
\begin{multline} \label{eq:proof:hybrid update operator:mknf:assumption:1}
\smod \left( \Set{ H^*(r) | r \in \progmod \land T_{\an{\ont, \progmod}}^n(\foint) \ent B(r) } \oplus^{\tbox} \abox \right) \\
= \smod \left( \ont \cup \Set{ H^*(r) | r \in \progmod \land T_{\an{\ont, \progmod}}^n(\foint) \ent B(r) } \right)
\end{multline}
By induction on $n$ we will now prove that $T_{\an{\ont, \progmod}}^n(\foint) = T_{\progmod \oplus^{\tbox} \abox}^n(\foint)$.
\renewcommand{\labelenumi}{\arabic{enumi}$^\circ$}
\begin{enumerate}
	\item For $n = 0$ we have
	\[
	T_{\an{\ont, \progmod}}^0(\foint) = \foint = T_{\progmod \oplus^{\tbox} \abox}^0(\foint)
	\]
	\item We assume the claim holds for $n-1$, i.e.
	\begin{equation} \label{eq:proof:hybrid update operator:mknf:inductive assumption:1}
	T_{\an{\ont, \progmod}}^{n-1}(\foint) = T_{\progmod \oplus^{\tbox} \abox}^{n-1}(\foint)
	\end{equation}
	and prove that it holds for $n$. Indeed, we obtain:
	\begin{align*}
	T_{\an{\ont, \progmod}}^n(\foint)
	&\eqbyeq
		\smod \left( \ont \cup \Set{ H^*(r) | r \in \progmod \land T_{\an{\ont, \progmod}}^{n-1}(\foint) \ent B(r) } \right) \\
	&\eqbyeq[\eqref{eq:proof:hybrid update operator:mknf:assumption:1}]
		\smod \left( \Set{ H^*(r) | r \in \progmod \land T_{\an{\ont	, \progmod}}^{n-1}(\foint) \ent B(r) } \oplus^{\tbox} \abox \right) \\
	&\eqbyeq[\eqref{eq:proof:hybrid update operator:mknf:inductive assumption:1}]
		\smod \left( \Set{ H^*(r) | r \in \progmod \land T_{\progmod \oplus^{\tbox} \abox}^{n-1}(\foint) \ent B(r) } \oplus^{\tbox} \abox \right) \\
	&\eqbyeq T_{\progmod \oplus^{\tbox} \abox}^n(\foint)
	\end{align*}
\end{enumerate}
So \eqref{eq:proof:hybrid update operator:mknf:dynamic model} is satisfied and consequently $M$ is a minimal change dynamic stable model of $\prog \oplus^{\tbox} \abox$.

For the converse statement, suppose $M$ is a minimal change dynamic stable model of $\prog \oplus^{\tbox} \abox$. Then from \eqref{eq:proof:hybrid update operator:mknf:mknf model} and Lemma \ref{lemma:mknf:mk_monotony} we obtain for every $n \in \nat$ that
\[
\Set{ H^*(r) | r \in \progmod \land T_{\progmod \oplus^{\tbox} \abox}^n(\foint) \ent B(r) } \subseteq \Set{ H^*(r) | r \in \prog \land M \ent B(r) } \enspace.
\]
Hence, by the assumption of the proposition,
\begin{multline} \label{eq:proof:hybrid update operator:mknf:assumption:2}
\smod \left( \Set{ H^*(r) | r \in \progmod \land T_{\progmod \oplus^{\tbox} \abox}^n(\foint) \ent B(r) } \oplus^{\tbox} \abox \right) \\
= \smod \left( \ont \cup \Set{ H^*(r) | r \in \progmod \land T_{\progmod \oplus^{\tbox} \abox}^n(\foint) \ent B(r) } \right)
\end{multline}
By induction on $n$ we will now prove that $T_{\progmod \oplus^{\tbox} \abox}^n(\foint) = T_{\an{\ont, \progmod}}^n(\foint)$.
\renewcommand{\labelenumi}{\arabic{enumi}$^\circ$}
\begin{enumerate}
	\item For $n = 0$ we have
	\[
	T_{\progmod \oplus^{\tbox} \abox}^0(\foint) = \foint = T_{\an{\ont, \progmod}}^0(\foint)
	\]
	\item We assume the claim holds for $n-1$, i.e.
	\begin{equation} \label{eq:proof:hybrid update operator:mknf:inductive assumption:2}
	T_{\progmod \oplus^{\tbox} \abox}^{n-1}(\foint) = T_{\an{\ont, \progmod}}^{n-1}(\foint)
	\end{equation}
	and prove that it holds for $n$. Indeed, we obtain:
	\begin{align*}
	T_{\progmod \oplus^{\tbox} \abox}^n(\foint)
	&\eqbyeq
		\smod \left( \Set{ H^*(r) | r \in \progmod \land T_{\progmod \oplus^{\tbox} \abox}^{n-1}(\foint) \ent B(r) } \oplus^{\tbox} \abox \right) \\
	&\eqbyeq[\eqref{eq:proof:hybrid update operator:mknf:assumption:2}]
		\smod \left( \ont \cup \Set{ H^*(r) | r \in \progmod \land T_{\progmod \oplus^{\tbox} \abox}^{n-1}(\foint) \ent B(r) } \right) \\
	&\eqbyeq[\eqref{eq:proof:hybrid update operator:mknf:inductive assumption:2}]
		\smod \left( \ont \cup \Set{ H^*(r) | r \in \progmod \land T_{\an{\ont, \progmod}}^{n-1}(\foint) \ent B(r) } \right) \\
	&\eqbyeq T_{\an{\ont, \progmod}}^n(\foint)
	\end{align*}
\end{enumerate}
So \eqref{eq:proof:hybrid update operator:mknf:mknf model} is satisfied and consequently $M$ is an MKNF model of $\an{\ont, \prog}$.
\end{proof}

\begin{corollary*}{cor:hybrid update operator:mknf:sm}
Let $\prog$ be a finite ground program. Then $M$ is a stable model of $\prog$ if and only if $M$ is a minimal change dynamic stable model of $\prog \oplus^{\emptyset} \emptyset$.
\end{corollary*}
\begin{proof}[Proof of Corollary \ref{cor:hybrid update operator:mknf:sm}]
\label{proof:hybrid update operator:mknf:sm}
Follows from the previous corollary and the fact that MKNF models coincide with stable models on the class of normal logic programs \cite{Lifschitz1991}.
\end{proof}

\begin{proposition*}{prop:hybrid update operator:generalisation of classical updates}
Let $\prog$ be a finite ground program containing only facts, $\tbox$ a TBox, $\abox$ a sequence of ABoxes and $M$ an MKNF interpretation. Then $M$ is a minimal change dynamic stable model of $\prog \oplus^{\tbox} \abox$ if and only if $M$ is a minimal change update model of $\thr_{\prog} \oplus^{\tbox} \abox$ where $\thr_{\prog} = \set{ p | \mk p \in \prog }$.
\end{proposition*}
\begin{proof}[Proof of Proposition \ref{prop:hybrid update operator:generalisation of classical updates}]
\label{proof:hybrid update operator:generalisation of classical updates}
Since $\prog$ contains only facts, we can see that $\prog = \progmod$, so $M$ is a minimal change dynamic stable model of $\prog \oplus^{\tbox} \abox$ if and only if $M = \smod(\prog \oplus^{\tbox} \abox)$ which by definition holds if and only if
\[
M = \bigcap_{n \geq 0} T_{\prog \oplus^{\tbox} \abox}^n( \foint )
\]
Further, we know that
\begin{align*}
T_{\prog \oplus^{\tbox} \abox}^0(\foint) &= \foint \\
T_{\prog \oplus^{\tbox} \abox}^1(\foint) &= \smod( \Set{ H^*(r) | r \in \prog \land \foint \ent B(r) } \oplus^{\tbox} \abox ) \\
	&= \smod( \Set{ H^*(r) | r \in \prog } \oplus^{\tbox} \abox ) = \smod( \thr_{\prog} \oplus^{\tbox} \abox ) \\
T_{\prog \oplus^{\tbox} \abox}^n(\foint) &= T_{\prog \oplus^{\tbox} \abox}^1(\foint) \quad \text{ for all } n > 1
\end{align*}
Hence, we have
\[
\bigcap_{n \geq 0} T_{\prog \oplus^{\tbox} \abox}^n( \foint ) = \smod(\thr_{\prog} \oplus^{\tbox} \abox) \enspace.
\]
So $M$ is a minimal change dynamic stable model of $P \oplus U$ if and only if $M = \smod(\thr_{\prog} \oplus^{\tbox} \abox)$ which is by definition equivalent to $M$ being a minimal change update model of $\thr_{\prog} \oplus^{\tbox} \abox$.
\end{proof}

\begin{lemma} \label{lemma:hybrid update operator:lemma empty theory}
Let $\tbox$ be a TBox, $\abox = (\abox_1, \abox_2, \dotsc, \abox_n)$ a sequence of ABoxes (where $n \geq 1$) and
\[
\abox' = (\abox_1, \abox_2, \dotsc, \abox_{i-1}, \abox_i, \emptyset, \abox_{i+1}, \dotsc, \abox_n)
\]
for some $i \in \set{0, 1, 2, \dotsc, n}$. Then for any $M \subseteq \smod(\tbox)$ it holds that
\[
\mathsf{incorporate}^{\tbox}(\abox, M) = \mathsf{incorporate}^{\tbox}(\abox', M)
\]
\end{lemma}
\begin{proof*}
We will prove by induction on $n$:
\renewcommand{\labelenumi}{\arabic{enumi}$^\circ$}
\begin{enumerate}
	\item If $n = 1$, then $i \in \set{0, 1}$, so we need to prove that
	\[
	\mathsf{incorporate}^{\tbox}(\abox_1, M) = \mathsf{incorporate}^{\tbox}(\abox_1, \mathsf{incorporate}^{\tbox}(\emptyset, M))
	\]
	and that
	\[
	\mathsf{incorporate}^{\tbox}(\abox_1, M) = \mathsf{incorporate}^{\tbox}(\emptyset, \mathsf{incorporate}^{\tbox}(\abox_1, M)) \enspace.
	\]
	This follows easily from the fact that $\mathsf{incorporate}^{\tbox}(\emptyset, N) = N$ for any $N \subseteq \smod(\tbox)$.
	
	\item We assume the claim holds for $n-1$ and prove it for $n$. First let $i = 0$. Then
	\[
	\mathsf{incorporate}^{\tbox}(\abox', M) = \mathsf{incorporate}^{\tbox}(\abox, \mathsf{incorporate}^{\tbox}(\emptyset, M))
	\]
	and the claim again follows from the fact that $\mathsf{incorporate}(\emptyset, N) = N$ for any $N \subseteq \smod(\tbox)$.

	Now suppose $i > 0$ and let
	\begin{align*}
	\abox[B] &= (\abox_2, \abox_3, \dotsc, \abox_n) \\
	\abox[B]' &= (\abox_2, \abox_3, \dotsc, \abox_{i-1}, \abox_i, \emptyset, \abox_{i+1}, \dotsc, \abox_n)
	\end{align*}
	By the inductive assumption we know that for any $N \subseteq \tbox$ it is holds that
	\[
	\mathsf{incorporate}^{\tbox}(\abox[B], N) = \mathsf{incorporate}^{\tbox}(\abox[B]', N)
	\]
	Hence,
	\begin{align*}
	\mathsf{incorporate}^{\tbox}(\abox', M) &= \mathsf{incorporate}^{\tbox}(\abox[B]', \mathsf{incorporate}^{\tbox}(\abox_1, M)) \\
	&= \mathsf{incorporate}^{\tbox}(\abox[B], \mathsf{incorporate}^{\tbox}(\abox_1, M)) \\
	& = \mathsf{incorporate}^{\tbox}(\abox, M) \enspace. \qedhere
	\end{align*}
\end{enumerate}
\end{proof*}

\begin{corollary} \label{cor:hybrid update operator:lemma empty theory}
Let $\tbox$ be a TBox, $\abox = (\abox_1, \abox_2, \dotsc, \abox_n)$ a sequence of ABoxes (where $n \geq 1$) and
\[
\abox' = (\abox_1, \abox_2, \dotsc, \abox_{i-1}, \abox_i, \emptyset, \abox_{i+1}, \dotsc, \abox_n)
\]
for some $i \in \set{0, 1, 2, \dotsc, n}$. Then for any first-order theory $\thr$ it holds that
\[
\smod(\thr \oplus^{\tbox} \abox) = \smod(\thr \oplus^{\tbox} \abox')
\]
\end{corollary}
\begin{proof}
Follows by applying the previous lemma to $M = \smod(\tbox \cup \thr)$.
\end{proof}

\begin{proposition*}{prop:hybrid update operator:empty theory}
Let $\prog$ be a finite ground program, $\tbox$ a TBox and $\abox$ a sequence of ABoxes  = $(\abox_1, \abox_2, \dotsc, \abox_n)$ (where $n \geq 1$). Let
\[
\abox' = (\abox_1, \abox_2, \dotsc, \abox_{i-1}, \abox_i, \emptyset, \abox_{i+1}, \dotsc, \abox_n)
\]
for some $i \in \set{0, 1, 2, \dotsc, n}$. Then an MKNF interpretation $M$ is a minimal change dynamic stable model of $\prog \oplus^{\tbox} \abox$ if and only if $M$ is a minimal change dynamic stable model of $\prog \oplus^{\tbox} \abox'$.
\end{proposition*}
\begin{proof*}[Proof of Proposition \ref{prop:hybrid update operator:empty theory}]
\label{proof:hybrid update operator:empty theory}
We need to show that
\[
\bigcap_{n \geq 0} T_{\progmod \oplus^{\tbox} \abox}^n(\foint) = \bigcap_{n \geq 0} T_{\progmod \oplus^{\tbox} \abox'}^n(\foint) \enspace.
\]
By induction on $n$ we will prove that for all $n \in \nat$ it holds that
\[
T_{\progmod \oplus^{\tbox} \abox}^n(\foint) = T_{\progmod \oplus^{\tbox} \abox'}^n(\foint) \enspace.
\]
\renewcommand{\labelenumi}{\arabic{enumi}$^\circ$}
\begin{enumerate}
	\item For $n = 0$ we directly obtain
	\[
	T_{\progmod \oplus^{\tbox} \abox}^n(\foint) = \foint = T_{\progmod \oplus^{\tbox} \abox'}^n(\foint) \enspace.
	\]
	\item We assume the claim holds for $n-1$ and prove it for $n$. We have
	\[
	T_{\progmod \oplus^{\tbox} \abox}^n(\foint) = \smod \left( \Set{ H^*(r) | r \in \progmod \land T_{\progmod \oplus^{\tbox} \abox}^{n-1}(\foint) \ent B(r) } \oplus^{\tbox} \abox \right) \enspace.
	\]
	By the inductive assumption we obtain that $T_{\progmod \oplus^{\tbox} \abox}^{n-1}(\foint) = T_{\progmod \oplus^{\tbox} \abox'}^{n-1}(\foint)$, so
	\[
	T_{\progmod \oplus^{\tbox} \abox}^n(\foint) = \smod \left( \Set{ H^*(r) | r \in \progmod \land T_{\progmod \oplus^{\tbox} \abox'}^{n-1}(\foint) \ent B(r) } \oplus^{\tbox} \abox \right) \enspace.
	\]
	Corollary \ref{cor:hybrid update operator:lemma empty theory} now implies that
	\begin{align*}
	T_{\progmod \oplus^{\tbox} \abox}^n(\foint) &= \smod \left( \Set{ H^*(r) | r \in \progmod \land T_{\progmod \oplus^{\tbox} \abox'}^{n-1}(\foint) \ent B(r) } \oplus^{\tbox} \abox' \right) \\
	&= T_{\progmod \oplus^{\tbox} \abox'}^n(\foint) \enspace. \qedhere
	\end{align*}
\end{enumerate}
\end{proof*}

\begin{proposition*}{prop:hybrid update operator:empty program}
Let $\tbox$ be a TBox, $\abox$ an ABox and $M$ an MKNF interpretation. Then $M$ is a minimal change dynamic stable model of $\emptyset \oplus^{\tbox} \abox$ if and only if $M = \smod(\tbox \cup \abox)$.
\end{proposition*}
\begin{proof}[Proof of Proposition \ref{prop:hybrid update operator:empty program}]
\label{proof:hybrid update operator:empty program}
By Proposition \ref{prop:hybrid update operator:tpu continuous} and Theorem \ref{thm:order:kleene}, $M$ is a minimal change dynamic stable model of $\emptyset \oplus^{\tbox} \abox$ if and only if
\[
M = \bigcap_{n \geq 0} T_{\progmod \oplus^{\tbox} \abox}^n( \foint )
\]
where
\begin{align*}
T_{\progmod \oplus^{\tbox} \abox}^0( \foint ) &= \foint \\
T_{\progmod \oplus^{\tbox} \abox}^1( \foint ) &= \smod( \Set{H^*(r) | r \in \progmod \land \foint \ent B(r) } \oplus^{\tbox} \abox ) = \smod( \emptyset \oplus^{\tbox} \abox ) \\
T_{\progmod \oplus^{\tbox} \abox}^n( \foint ) &= T_{\progmod \oplus^{\tbox} \abox}^1( \foint ) \quad \text{ for all } n > 1
\end{align*}
So $M$ is a minimal change dynamic stable model of $\emptyset \oplus^{\tbox} \abox$ if and only if $M = \smod(\emptyset \oplus^{\tbox} \abox)$. Further,
\[
\smod(\emptyset \oplus^{\tbox} \abox) = \mathsf{incorporate}^{\tbox}(\abox, \smod(\emptyset)) = \mathsf{incorporate}^{\tbox}(\abox, \foint) = \smod(\tbox \cup \abox) \enspace.
\]
Hence, $M$ is a minimal change dynamic stable model  of $\emptyset \oplus^{\tbox} \abox$ if and only if $M = \smod(\tbox \cup \abox)$.
\end{proof}

\end{document}